\documentclass[11pt]{article}
\usepackage[T1]{fontenc}
\usepackage[utf8]{inputenc}
\usepackage{roboto}
\usepackage{newtxtext}
\usepackage{newtxmath}

\usepackage[letterpaper,margin=1in]{geometry}
\usepackage{microtype}
\usepackage{graphicx}
\usepackage{booktabs}
\usepackage{tabularx}
\usepackage{multirow}
\usepackage{xcolor}
\usepackage{enumitem}
\usepackage{amsmath}
\usepackage{caption}
\usepackage[numbers,sort&compress]{natbib}
\usepackage{titlesec}
\usepackage{fancyhdr}
\usepackage{tcolorbox}
\usepackage[section]{placeins}

\usepackage[colorlinks=true,linkcolor=accent,citecolor=accent,urlcolor=accent]{hyperref}
\usepackage[nameinlink,capitalise]{cleveref}

\definecolor{accent}{HTML}{5B3DF0}
\definecolor{ink2}{HTML}{52514E}
\definecolor{wash}{HTML}{F4F2FF}

\titleformat{\section}{\sffamily\Large\bfseries}{\textcolor{accent}{\thesection}}{0.7em}{}
\titleformat{\subsection}{\sffamily\large\bfseries}{\textcolor{accent}{\thesubsection}}{0.6em}{}
\titleformat{\paragraph}[runin]{\sffamily\bfseries}{}{}{}[.]
\titlespacing*{\section}{0pt}{2.0ex plus .4ex}{1.0ex}
\titlespacing*{\subsection}{0pt}{1.6ex plus .3ex}{0.7ex}
\setlist{itemsep=2pt,topsep=4pt,leftmargin=1.4em}
\newcommand{\twin}{TwinStore}
\newtcolorbox{hlbox}{colback=wash,colframe=wash,arc=6pt,boxrule=0pt,left=10pt,right=10pt,top=7pt,bottom=7pt,fontupper=\small}

\begin{document}
\thispagestyle{empty}
\begin{center}
{\sffamily\bfseries\LARGE TwinStore: A Synthetic Consumer Platform\\[2pt] Using Vision--Language Models\par}\vspace{8pt}
{\sffamily\small\color{ink2}Technical Report\par}\vspace{6pt}
{\sffamily\large Minhyeok Lee\par}\vspace{3pt}
{\sffamily\small\color{ink2}\url{https://github.com/BrainJellyPie/twinstore} \quad·\quad October 2026\par}
\end{center}
\vspace{-6pt}
\begin{abstract}
\noindent We present \twin{}, a source-available synthetic consumer platform for physical retail built on vision--language models. \twin{} turns a store into a digital twin---from 82 ready-made formats, a floor-plan image or phone photos of its shelves---and fills it with synthetic shoppers who plan a trip, walk the aisles and decide what to buy by \emph{looking} at rendered shelves with packs, price tags and promotion signs. Every trip ends in a receipt, so a layout, price, pack or promotion can be tried on the same shoppers before it reaches the store. Around this core, \twin{} provides 3D playback, paired comparisons and layout optimization, shopper interviews, a trade-area model, counterfactual edits of real shelf photos, a marketing studio with eleven tools, and an assistant that turns a plain sentence into a drafted study or an answer grounded in a run's own numbers. Every model call is seeded and cached, so runs are reproducible and comparisons are paired. In a held-out validation on eleven public datasets, simulated receipts approach the households' own natural variation (category-mix divergence 0.017--0.018 against a real drift of 0.012), price elasticities fall inside the range measured in the field, and three independent decision models agree. \twin{} runs on cloud or on-device models and is distributed with installers for Windows, macOS and Linux.
\end{abstract}
\vspace{2pt}
\noindent\begin{minipage}{\linewidth}\centering
  \includegraphics[width=0.83\linewidth]{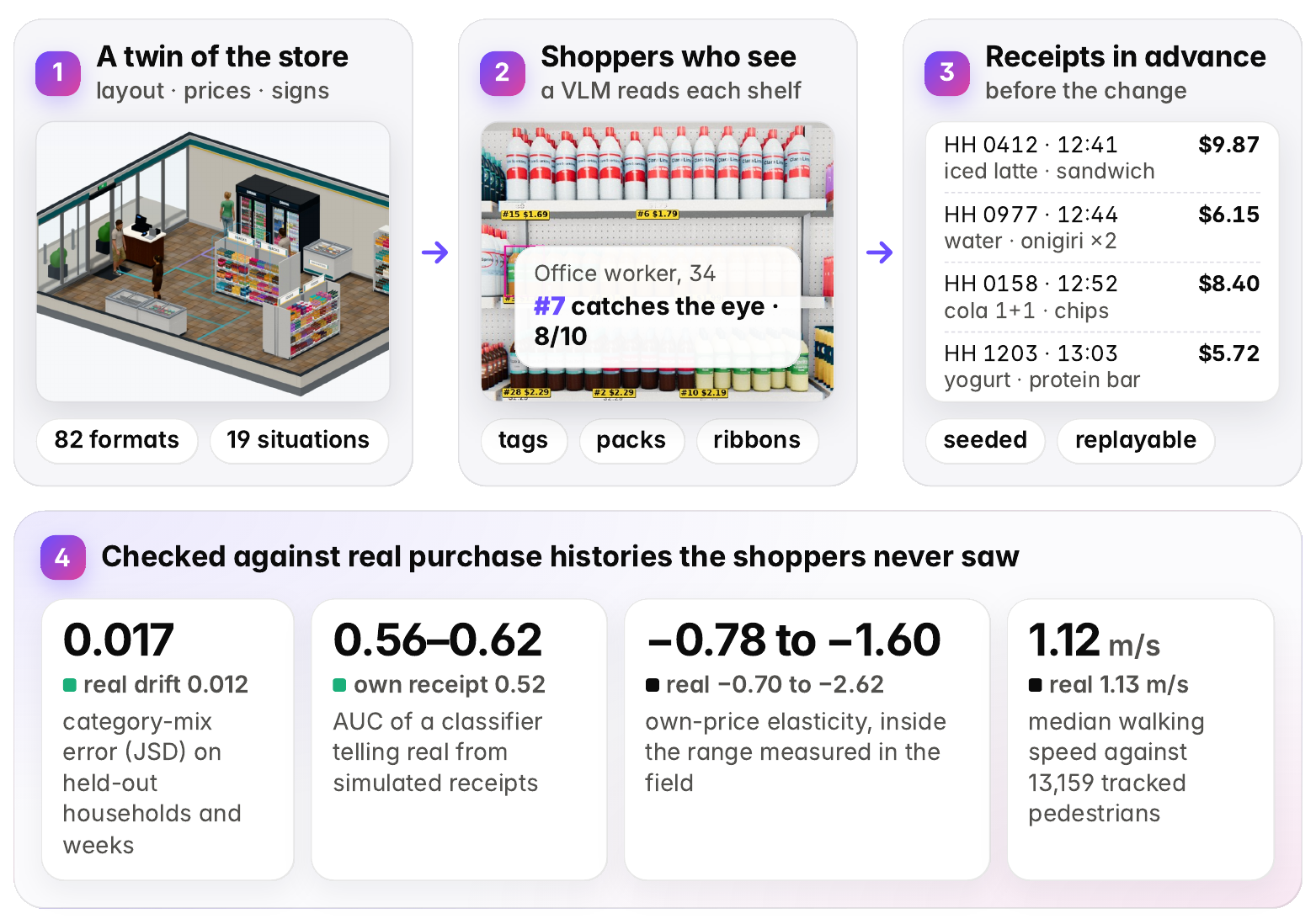}
  \captionof{figure}{\textbf{\twin{} at a glance.} (1) A digital twin holds the store's layout, assortment, prices and signs. (2) Synthetic shoppers decide by looking at rendered shelves, as a person in the aisle would. (3) Their trips become receipts for a store that has not changed yet. (4) Against real purchase histories they never saw, these receipts come close to the households' own natural variation.}
  \label{fig:teaser}
\end{minipage}
\clearpage

\section{Introduction}

Retail decisions are judged by the purchase history they produce: a new layout, a price cut, a redesigned pack or a promotion is good if the receipts that follow are better. That history arrives only after the decision has been made. Field experiments measure it directly but take weeks, test one or two alternatives at a time and are confounded by weather, competitors and seasonality. Demand models extrapolate from the variation that history happened to contain and have no handle on what shoppers see at the shelf---the pack, the facing, the tag---which drives much of in-store choice~\citep{inman2009,chandon2009}. Surveys and conjoint studies are fast, but they measure stated preferences.

Language models conditioned on a description of a person reproduce the answers of real survey subgroups~\citep{argyle2023}, classic behavioral experiments~\citep{horton2023,aher2023} and willingness-to-pay estimates of human panels~\citep{brand2023}, and agent architectures let simulated people act coherently over time~\citep{park2023,park2024}. These synthetic consumers read text, while the point of purchase is visual, and they have been validated mainly against stated responses rather than against the purchase logs a retailer earns money from.

\twin{} is a platform built to close both gaps. It combines a digital twin of the store with synthetic shoppers whose decisions are made by vision--language models looking at the shelves, records every trip as a receipt, and wraps this core in the tools that store operators, brand marketers and research agencies use to make decisions. This report describes the platform: its architecture (\cref{sec:overview}), how a twin is built (\cref{sec:twin}), how shoppers decide (\cref{sec:shoppers}), the tools built on them (\cref{sec:tools}), a held-out validation against real purchase data (\cref{sec:validation}), implementation and deployment (\cref{sec:impl}), related work, limitations and licensing.

\begin{hlbox}
\textbf{\sffamily Contributions}
\begin{itemize}[leftmargin=1.2em,itemsep=1pt,topsep=3pt]
  \item A vision-based shopper model that renders each shelf face a shopper passes---packs, price tags and promotion signs---and lets a vision--language model choose what draws attention, so that changes to visual merchandising reach the purchase decision.
  \item Store construction from 82 parameterized formats and 25 multi-floor buildings, from floor-plan images and from shelf photographs.
  \item Paired, reproducible simulation: per-shopper random streams and cached model calls make runs replayable and comparisons between store configurations paired.
  \item An integrated set of tools on the same simulated shoppers: playback, paired comparison, layout optimization, journeys and interviews, a trade-area model, counterfactual edits of shelf photographs and eleven research tools, with a language-model assistant that drafts studies from a stated question and answers questions from a run's own statistics.
  \item A held-out validation against revealed purchases on eleven public datasets from six countries, replicated across three decision models.
\end{itemize}
\end{hlbox}

\section{Design goals}
Five goals shaped \twin{}.
\begin{enumerate}[label=\textbf{G\arabic*}]
  \item \textbf{Decide from what is seen.} Many retail levers act only through perception. A shopper whose input is a product list cannot respond to a new pack, an extra facing or a promotion ribbon; a shopper who sees the shelf can.
  \item \textbf{Compare like with like.} A difference between two stores must come from the change alone. The same shoppers, the same random draws and the same unaffected decisions are reused across alternatives.
  \item \textbf{From the real store to its twin quickly.} Building a store by hand is the main cost of simulation. Templates, floor-plan reading and shelf-photo reading reduce it to minutes.
  \item \textbf{Usable without a programmer.} Installation, model setup, experiments and reports all happen in a graphical interface.
  \item \textbf{Checked against reality.} The shoppers are tested on real purchase data they have not seen, and their known limits are stated.
\end{enumerate}

\section{System overview}\label{sec:overview}

\begin{figure}[!tbp]
  \centering
  \includegraphics[width=\linewidth]{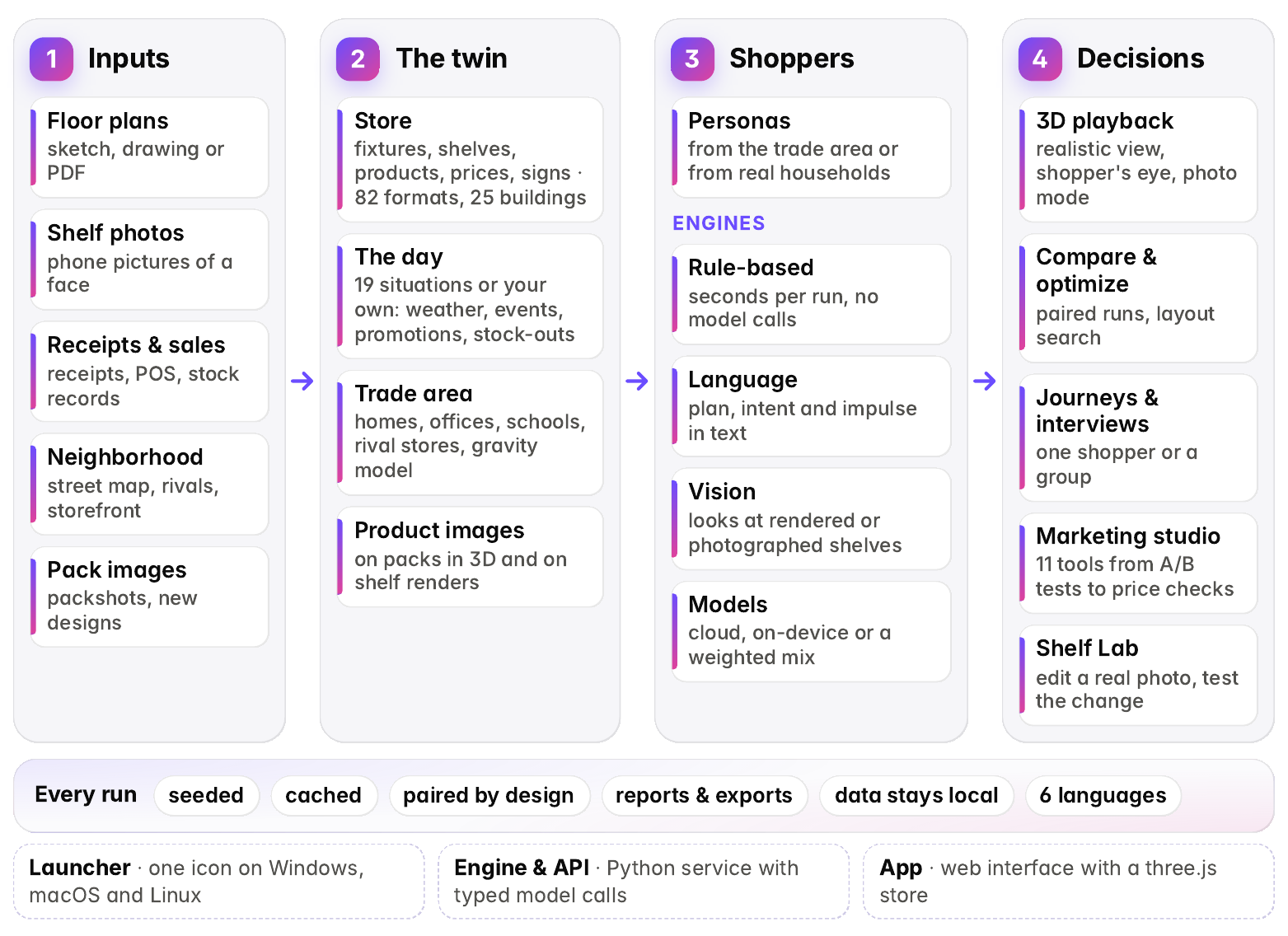}
  \caption{\textbf{Architecture.} Inputs from the real store (1) build the twin: the store itself, the situation of the day and the trade area (2). Synthetic shoppers run on one of three engines and on cloud, on-device or mixed models (3). Their trips feed the decision tools (4). Every run is seeded, cached and paired; the platform ships as a launcher, an engine with an API, and a web application.}
  \label{fig:arch}
\end{figure}

\Cref{fig:arch} shows the platform. Inputs from the real store---floor plans, shelf photos, receipts and sales, the neighborhood and pack images---build a twin made of the store, the situation of the day and the trade area. Synthetic shoppers generated for the store or grounded in real households shop the twin on one of three engines. Their trips feed the decision tools.

\paragraph{Software} \twin{} consists of three parts. A Python engine holds the domain model, the simulation engines and the analyses, and calls language and vision models through a provider-independent layer with validated, structured outputs. A web application renders the store in 3D and hosts every tool, and a launcher installs and starts both on Windows, macOS and Linux. Projects are stored as files on the user's machine; nothing leaves it except the model calls the user configures.

\paragraph{Relation to Parastore} \twin{} extends Parastore~\citep{parastore}, an open-source store simulator released by intellicia under the MIT License. The vision and rule-based engines, store generation from formats, floor plans and shelf photographs, situations, the trade-area model, the analysis and interview tools, the marketing studio, Shelf Lab, the assistant, local models and the launcher are new in \twin{}. The grid store model, persona and arrival generation, path planning and footfall maps, the three-stage language engine and the 3D playback originate in Parastore and were extended in \twin{}.

\section{Building the twin}\label{sec:twin}

\subsection{The store model}
A store is a grid of fixtures---shelves, fridges, freezers, counters, tables and checkouts---with entrances and exits on its walls. Each fixture has faces, levels and slots; each slot holds a product with a name, a category, a price and optional promotions: a discount shown with its former price, a ``1+1'' or ``SALE'' ribbon, or a move to eye level. Opening hours and open weekdays are kept per store and per floor. Product images uploaded by the user are placed on the packs in the 3D store and on the shelf images the shoppers see.

\subsection{Formats and buildings}
\twin{} ships 82 furnished store formats---convenience stores, supermarkets, hypermarkets and warehouse clubs, pharmacies, cafés, bakeries and restaurants, apparel, footwear and eyewear, bookstores, electronics, pet and garden stores, refill grocers, fragrance boutiques, bubble tea cafés, Indian, Vietnamese, Thai, Turkish, Mediterranean and Mexican restaurants, airport and hospital shops, pet and outdoor superstores and more---and 25 multi-floor buildings such as a department store, a campus hub, a dining hall and a hospital services center. Each format comes with its own assortment, interior design, seating and checkout count. A generated store can be adjusted before it is created: the product categories and the number of products per category, the walls of the entrance, exit and checkout, aisle width and shelf direction, up to 48 dining tables of two, four and six seats in exact numbers with their orientation and spacing, reserved areas, the interior design (from Nordic to retro, or the store's own colors), opening hours for each weekday, and a \emph{layout priority} that sets several of these at once (eight presets, from wide circulation and more display space to family dining and separated service counters). Assortment, layout priorities and opening hours can be applied to every floor of a building in one step. A preview shows the floor plan and checks capacity, and the result can be compared with the format's default.

\subsection{From floor plans and shelf photos}
A floor plan---a hand sketch, an architectural drawing or a PDF---is read into walls, fixtures and doors and becomes the store's grid. A phone photo of a shelf face is read into shelves, product slots, product names and price tags, producing a planogram whose entries can be corrected in a list, with empty slots and facing counts shown. Several photos can be queued at once, and the reviewed photos of many shelves are combined into one layout; the names, prices and promotion texts of placed products can be searched and edited in bulk and saved as a new layout. The same captures power Shelf Lab (\cref{sec:shelflab}) and let the vision shoppers look at the real photo instead of a render.

\subsection{The situation of the day}
A situation describes the conditions of a simulated day: weekday and hour, weather and temperature, holidays and local events, crowding, stock-outs, promotions and notes for the shoppers. Nineteen are preset---from a rainy commute and a heatwave weekend to payday Friday night, a big match, a festival or a rival opening next door---and any combination can be composed.

\subsection{The trade area}
A neighborhood around the store is generated by type and density (residential, university, station, suburban, tourist or mixed) or imported from OpenStreetMap. Homes, workplaces, schools and rival stores are placed on the map, and visits are allocated by a gravity model~\citep{huff1964}. A photo of the storefront is scored for its signage, entrance and windows; photos of rival shelves are compared on price level and assortment overlap, with the same product matched even when it is named differently; and linked trips from nearby venues are taken into account. The resulting visitors can be sent into the store as the day's shoppers, each knowing where they came from.

\section{Synthetic shoppers}\label{sec:shoppers}

\subsection{Personas}
A shopper is a persona: age, gender, income, household composition, occupation, a budget, a reason for the visit and a mood. Personas are generated from the store's address, format and opening hours, from the trade area, or---in validation and for retailers with their own data---from real households' attributes and receipts, including their usual products and typical trip size. When the store has a trade area, generation draws on it: who comes on weekdays and at weekends, the busiest hours and how far people walk. Supplemental files such as customer research can be added to shape the population.

\subsection{Three engines}
\twin{} offers three engines that share the same store, personas, movement model and outputs (\cref{tab:engines}). The \emph{rule-based} engine decides with probabilities fitted to receipts and runs in seconds without model calls; it is suited to screening many alternatives. The \emph{language} engine, inherited from Parastore, decides in three stages from text---the shopping pattern, the planned purchases and the impulse purchases---and is suited to rich reasoning and interviews. The \emph{vision} engine, described next, decides from images.

\begin{table}[!tbp]
\centering\small
\caption{\textbf{The three shopper engines.}}
\label{tab:engines}
\begin{tabularx}{\linewidth}{@{}>{\sffamily\bfseries}l>{\raggedright\arraybackslash}X>{\raggedright\arraybackslash}X>{\raggedright\arraybackslash}X@{}}
\toprule
 & \sffamily Rule-based & \sffamily Language & \sffamily Vision\\
\midrule
Decides from & probabilities fitted to receipts & text: persona, store directory, situation & images: the store map at the door and rendered or photographed shelf faces\\
Plan & rule plan & pattern, then intent & one call at the entrance\\
Impulse purchases & probabilities & a text stage & glances at the best-seen faces\\
Reacts to packs, facings, signs & no & no & yes\\
Model calls per shopper & none & about three & one plan and one to three glances\\
Typical use & screening many options & reasoning, interviews & visual levers, final checks\\
\bottomrule
\end{tabularx}
\end{table}

\subsection{The vision shopper}

\begin{figure}[!tbp]
  \centering
  \includegraphics[width=\linewidth]{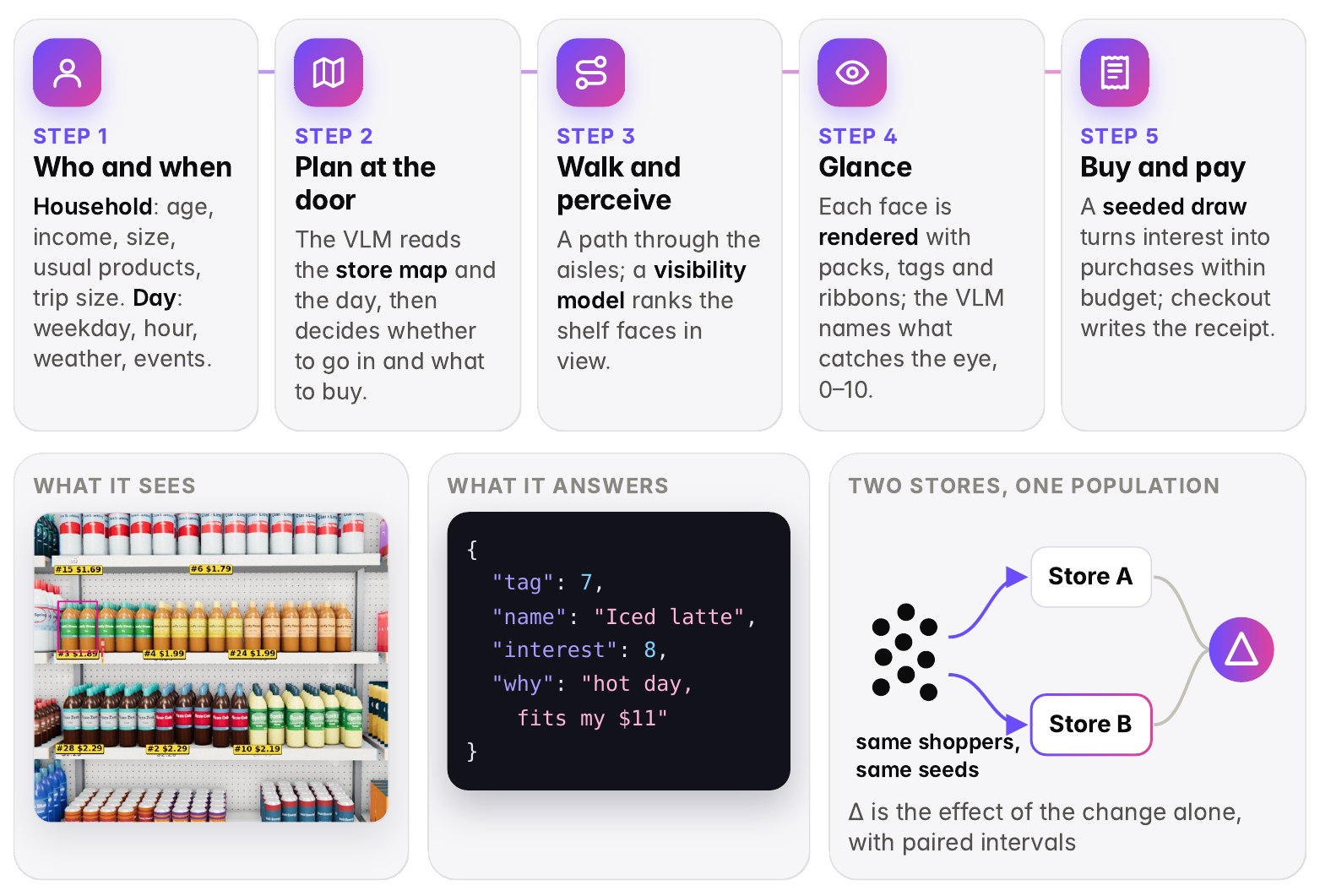}
  \caption{\textbf{One vision shopper's trip.} A household with its attributes and usual products meets the day's situation (1), plans at the entrance from the store map (2), walks a path along which a visibility model ranks the shelf faces in view (3), glances at rendered faces and names what catches the eye with an interest score (4), and buys through a seeded draw within budget (5). Below: the image the model receives, its structured answer, and the paired design that runs one population through two stores.}
  \label{fig:method}
\end{figure}

\Cref{fig:method} follows one trip. At the entrance, a single call to the vision--language model receives the persona, the hour, the situation, the money available, a text directory of the store and an image of its map; it decides whether the shopper goes in, the mission of the trip and a handful of planned items. Named items are matched to products on the shelves, with ties broken by the shopper's own random stream, so a shopper always takes the same product.

Along the route, a visibility model ranks the shelf faces by distance, angle and exposure time. For the best-seen faces the twin renders the face as an image: packs with their printed names, a yellow tag ``\#$N$ price'' under every product and promotion ribbons where they apply. The model, role-playing the persona with its mood, basket and remaining money, returns the number of the product that catches its eye, an interest rating from 0 to 10 and a reason. A seeded draw converts the rating into a purchase, a browse or nothing, within the budget and the household's trip size. The model decides \emph{what} tempts the shopper; one calibrated scale decides \emph{how often} temptation becomes a purchase. With a captured shelf, the shopper can look at the real photo, with numbered tags drawn on it, instead of a render. The mission and reason the model gives at the entrance become the first step of the shopper's journey and the speech bubble at the door; a shopper who turns away says why.

\subsection{Movement}
A path planner routes each shopper through the aisles between stops. Walking speed and time at the shelf are drawn from distributions matched to pedestrian tracking and eye-tracking measurements (\cref{sec:validation}). In the realistic view, lifelike 3D people walk with a gait matched to their speed, stop while they browse, pick up and pay, turn their heads and blink.

\subsection{Models}
Any OpenAI, Anthropic, Google, Qwen or OpenRouter model, or any OpenAI-compatible server such as vLLM, Ollama or llama.cpp, can drive the language and vision engines. Without a key, \twin{} selects an open vision--language model sized to the computer, downloads it and runs it locally. A run can also use a weighted mix of models, each shopper assigned one at random in proportion to its weight; mixed populations proved robust in validation.

\subsection{Determinism and paired designs}
Each shopper carries its own random streams, and each model response is cached by its full input, image included. Running one population through two stores therefore holds every random draw and every unaffected decision fixed, and the difference between the two receipt logs is the effect of the change, with variance reduced by common random numbers~\citep{glasserman1992}. Replicates with new seeds measure run-to-run noise, and paired bootstrap intervals accompany every comparison. Because responses are cached, a figure or report can be regenerated without new model calls.

\section{Tools for decisions}\label{sec:tools}

\begin{figure}[!tbp]
  \centering
  \includegraphics[width=\linewidth]{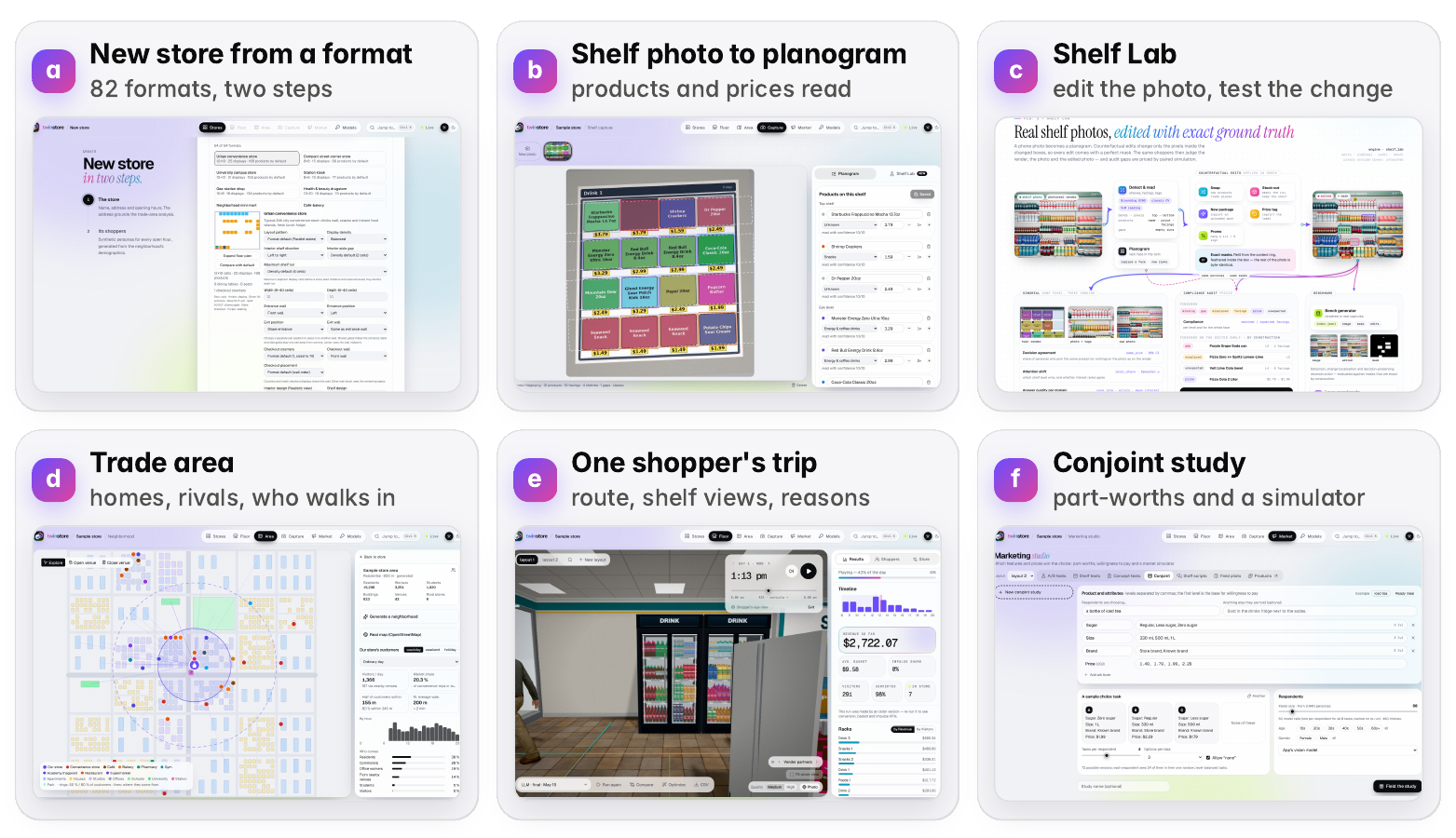}
  \caption{\textbf{A tour of the application.} (a) A new store from one of 82 formats in two steps. (b) A phone photo of a shelf read into a planogram. (c) Shelf Lab: counterfactual edits of a real shelf photo. (d) The trade area with homes, rival stores and the store's visitors. (e) One shopper's trip with the shelves they saw and the reasons they gave. (f) A choice-based conjoint study with its market simulator.}
  \label{fig:tour}
\end{figure}

\subsection{Playback and results}
A simulated day plays back in 3D (\cref{fig:tour}), in a light standard view or a realistic view with lifelike shoppers, from real time to fast-forward. Thoughts appear above the shoppers, any shopper can be followed or seen through their own eyes, and a photo mode renders the current view as a still with reflections and shadows. Results show revenue, basket size, conversion, impulse share and shoppers in the store by hour; floor maps of footfall, dwell and predicted congestion; fixture rankings; the shelf level at which products were picked; and integrity checks such as unprocessed shoppers or over-budget trips. Per-shopper results export to spreadsheets and can be matched against real receipts or sales.

\subsection{Comparison and optimization}
Two runs on the same shoppers are compared metric by metric, with paired differences, intervals and a one-line verdict. A layout optimizer searches for a better arrangement toward a chosen goal---revenue per visitor, conversion, units, impulse revenue or shorter paths---by moving products only or fixtures as well; the fast rule-based engine proposes candidates on thousands of shoppers, and vision shoppers judge the finalists on a paired panel. The original layout is kept and the best candidate is saved as a new one.

\subsection{Journeys and interviews}
The shopper list sorts the day's shoppers by spend, empty-handed exits, impulse share or time in store. A single shopper's trip shows the route and stops on the floor plan, the shelf each stop showed and the thought of that moment, and can be replayed in 3D from the moment the shopper entered. Any shopper can then be interviewed: the answer is grounded in the persona and the trip and cites the steps it relies on, which are highlighted on screen. A group interview puts one question to four to forty shoppers from a segment and groups the answers into themes.

\subsection{Shelf Lab}\label{sec:shelflab}
Shelf Lab edits a real shelf photo: two products swap places, a row empties to simulate a stock-out, a new pack is pasted in, a price tag changes or a promotion sign is added. Only the edited pixels change, so any difference in the shoppers' choices comes from the edit. The same shoppers can be shown the render, the tagged photo and the raw photo to compare decisions across them. Against a standard planogram, Shelf Lab finds missing and misplaced products, estimates the daily revenue they cost, ranks them for restocking, and turns stock, required facings, case sizes and unit costs into a restocking plan within a budget.

\subsection{Marketing studio}
The marketing studio gathers eleven tools on the same synthetic consumers (\cref{tab:studio}). Every finished study can be duplicated and rerun with new conditions, and results export as spreadsheets and reports.

\begin{table}[!tbp]
\centering\small
\caption{\textbf{The eleven tools of the marketing studio.}}
\label{tab:studio}
\begin{tabularx}{\linewidth}{@{}>{\sffamily\bfseries}l>{\raggedright\arraybackslash}X>{\raggedright\arraybackslash}X@{}}
\toprule
\sffamily Tool & \sffamily Question & \sffamily Output\\
\midrule
A/B tests & What do a price, discount, sign, placement or new pack change? & paired effects by segment with intervals; an automatic report; price sweeps with a demand curve, elasticity and the revenue-maximizing price\\
Shelf tests & Does a new pack, price or sign stand out on the shelf? & paired pick share and stand-out per respondent\\
Concept tests & Which pack or price do respondents prefer? & five-point ratings, top-two-box shares, preference shares, acceptable price ranges, verbatims\\
Conjoint & Which features and prices win the choice? & part-worths with intervals, importance, willingness to pay, a market simulator\\
Staff scripts & Which wording sells better? & paired purchase intent, trust and pressure\\
Field pilots & Does it hold in the real store? & a shift-balanced schedule and sample size; observed conversion; leave-one-out checks by staff and shift; staff time and a cost scenario\\
Protocols & What was planned, and what changed? & the original plan of a study with every later amendment and its reason\\
Price checks & Does the checkout charge what the shelf says? & matched label and checkout records with every difference and every excluded record\\
Unit prices & Are we dearer than nearby rivals? & observed offers converted to a price per kilogram, liter or item, with stale and conditional offers set aside\\
Inventory & Do the stock records add up? & reconciliation, period margin, stock turnover, net sell-through\\
Products & -- & pack images with backgrounds removed, linked to the products on the shelves\\
\bottomrule
\end{tabularx}
\end{table}

The conjoint tool, for example, gives each respondent a level-balanced random design of two to four profiles plus a ``none'' option, collects one choice per task, and fits a pooled multinomial logit with respondent-level bootstrap intervals; part-worths, attribute importance and willingness to pay follow, and the market simulator recomputes shares as products and prices change. Field pilots carry a promising script from the studio into the store: assignments are balanced across real shifts, encounters and checkout transactions are imported, and the observed conversion difference is judged once the planned sample is complete.

\subsection{Assistant}
A request typed into the command palette in plain language---``would a 1+1 sign on cola sell more than a 20\,\% cut?''---is routed by a language model to the tool that can answer it: an A/B test, a conjoint study, a staff script, the neighborhood, or a question about the last run. The tool opens with a draft built from the same words; nothing runs until the user has checked it. Drafting is also available inside each tool. The A/B builder turns a question into variants on the layout's own products and shelves, using the last run's sales to find the actual best seller; the conjoint builder proposes attributes, levels and price levels in the study's currency; ``rainy Friday evening, the game next door at 7, 1+1 on Pepsi, out of ice cream'' becomes a situation with its promotion and stock-out matched to the layout's products; a description of the area sets up the neighborhood; the new-store form writes the customer profile; and a staff-script test writes script B from script A. After a run, \emph{Ask about this run} answers questions from the run's own fact sheet and lists the figures each answer rests on, and a comparison of two runs is explained in terms of the shelves, products, hours and groups the difference came from, separating it from noise. Study results gain a model-written reading that groups respondents' reasons into themes, and interviews suggest follow-up questions fitted to the shopper's trip and the conversation.

\section{Validation}\label{sec:validation}

We tested whether the vision shoppers buy what real shoppers buy, react to the levers retailers pull as real shoppers do, and move as real people move, on public data they had not seen.

\begin{figure}[!tbp]
  \centering
  \includegraphics[width=\linewidth]{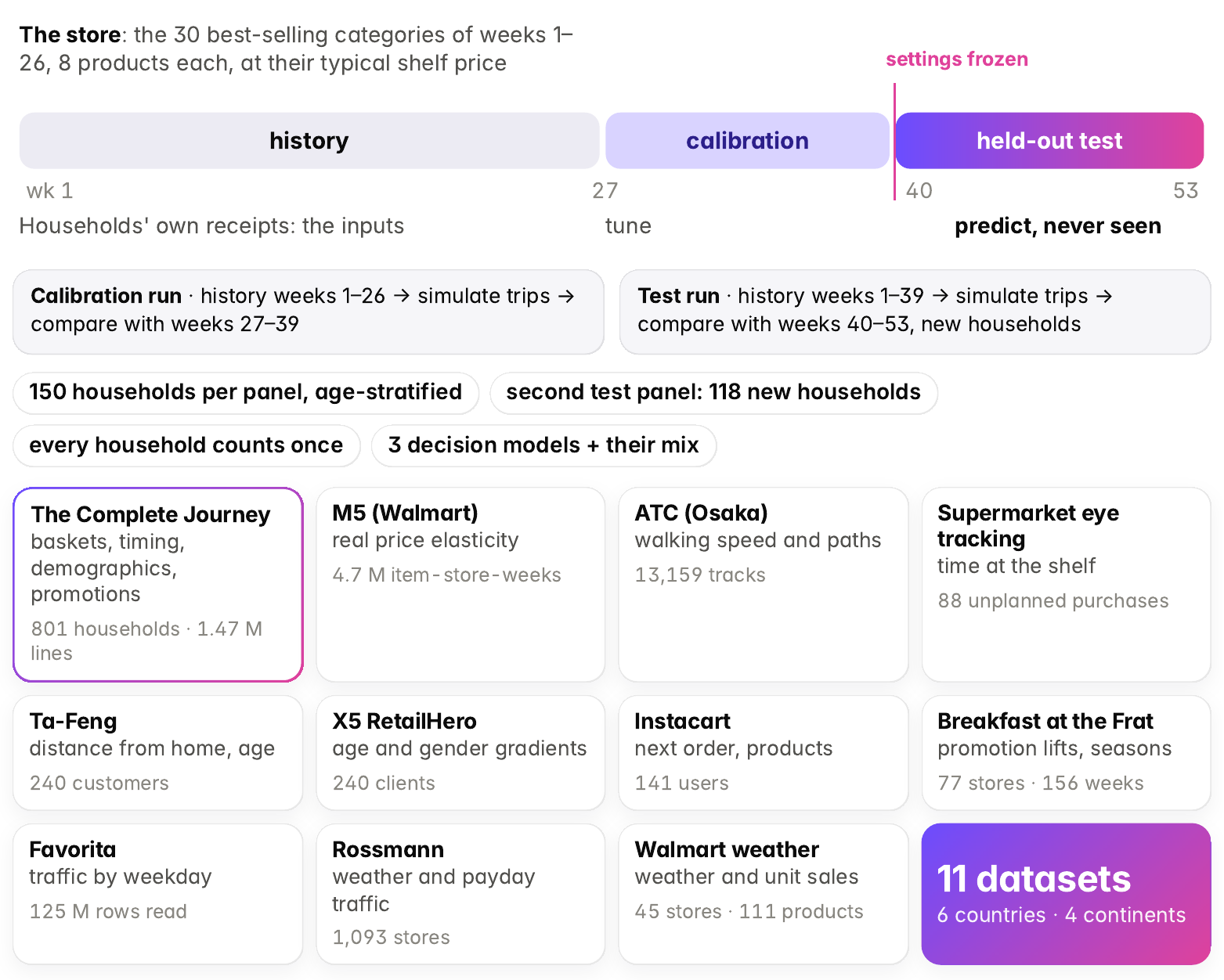}
  \caption{\textbf{Protocol and data.} Top: on \emph{The Complete Journey}, households' receipts of the history weeks are the only inputs; settings are tuned on weeks 27--39 and frozen; the test predicts weeks 40--53 for households excluded from calibration, with a second test panel of new households. Bottom: eleven public datasets test individual aspects of shopping behavior.}
  \label{fig:protocol}
\end{figure}

\paragraph{Protocol} The main test uses \emph{The Complete Journey}~\citep{cj2017}, the receipts, demographics and promotions of 801 households. The store holds the 30 best-selling categories of the first 26 weeks, eight products each at their typical price. Settings are chosen on a calibration window (history weeks 1--26, targets 27--39) and frozen; the test window (history 1--39, targets 40--53) uses 150 age-stratified households excluded from calibration, and a second test panel adds 118 new households (\cref{fig:protocol}). Ten further datasets test single aspects, including price elasticity (M5;~\citealp{makridakis2022}) and walking speed (ATC;~\citealp{brscic2013}). The vision engine runs with three decision models from three providers---Claude Haiku~5.5, Qwen~3.8 Flash and GPT-6 Luna---and with a population that mixes them. Every result is reported against a real reference, most often the households' own earlier receipts. The datasets are available from their providers under their own terms.

\begin{figure}[!tbp]
  \centering
  \includegraphics[width=0.84\linewidth]{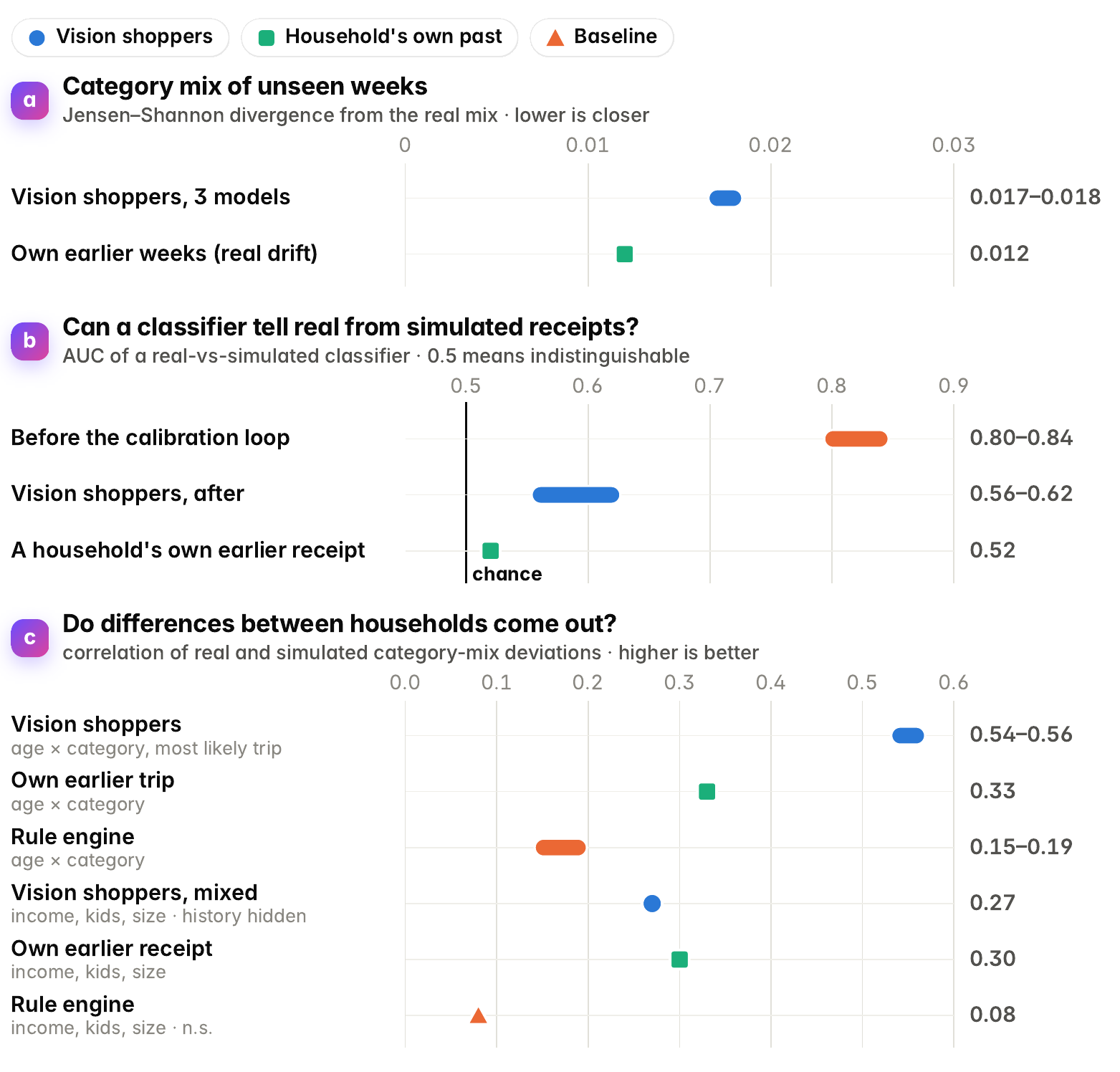}
  \caption{\textbf{As close as the households' own past.} (a) The category mix of simulated trips differs from the real mix of unseen weeks by little more than real households differ from their own earlier weeks. (b) A classifier separates simulated from real receipts only slightly better than a household's earlier from its later receipts. (c) Differences between households: by age, the most likely simulated trip tracks the real deviations more closely than one of the household's own earlier trips; by income, children and household size, shoppers who never see their history match one of their own past receipts. The rule engine does not.}
  \label{fig:fidelity}
\end{figure}

\paragraph{Buying the right things} On unseen households and weeks, the simulated category mix differs from the real one by a Jensen--Shannon divergence~\citep{lin1991} of 0.017--0.018; the households' own drift from their earlier weeks is 0.012 (\cref{fig:fidelity}a). Paired by household, every model's most likely trip predicts purchases better than the store's best-sellers ($+0.027$ to $+0.035$ in hit rate, 95\,\% intervals above zero). Told only who a household is and shown none of its history, the mixed population reproduces demographic differences in category mix with $r=0.27$ ($p=0.006$), against $0.30$ for one of the household's own earlier receipts and $0.08$ for the rule engine. Repeat purchases make up 56--72\,\% of a simulated trip, against 73\,\% in real trips.

\begin{table}[!htbp]
\centering\small
\caption{\textbf{Validation summary} on held-out data. Ranges span decision models and modes.}
\label{tab:evidence}
\begin{tabularx}{\linewidth}{@{}l>{\raggedright\arraybackslash}X>{\raggedright\arraybackslash}X@{}}
\toprule
\sffamily Check & \sffamily Vision shoppers & \sffamily Real reference\\
\midrule
Category mix (JSD) & 0.017--0.018 & 0.012 (own earlier weeks)\\
Household hit vs.\ best-sellers & $+0.027$ to $+0.035$, CI $>0$ & --\\
Demographic deviations, history hidden ($r$) & 0.27 ($p=0.006$) & 0.30 (own earlier receipt)\\
Repeat-purchase share & 0.56--0.72 & 0.73\\
Real vs.\ simulated classifier (AUC) & 0.56--0.62 & 0.52 (own earlier receipt)\\
Own-price elasticity & $-0.78$ to $-1.60$ & $-0.70$ to $-2.62$\\
Promotion lift & 1.25--1.7$\times$ & 1.4$\times$ (display)\\
Trip size, far vs.\ near & 1.25--1.36$\times$ & 1.33$\times$\\
Effects of the day ($r$) & 0.75--0.87 & --\\
Walking speed, median & 1.12\,m/s & 1.13\,m/s\\
Time at the shelf per impulse buy & 7--9\,s & 8.2\,s\\
\bottomrule
\end{tabularx}
\end{table}

\paragraph{Telling real from simulated} Before calibration, a classifier separated real from simulated receipts with an AUC of 0.80--0.84; after it, 0.56--0.62, against 0.52 for a household's own earlier versus later receipt (\cref{fig:fidelity}b)~\citep{lopezpaz2017}.

\begin{figure}[!tbp]
  \centering
  \includegraphics[width=\linewidth]{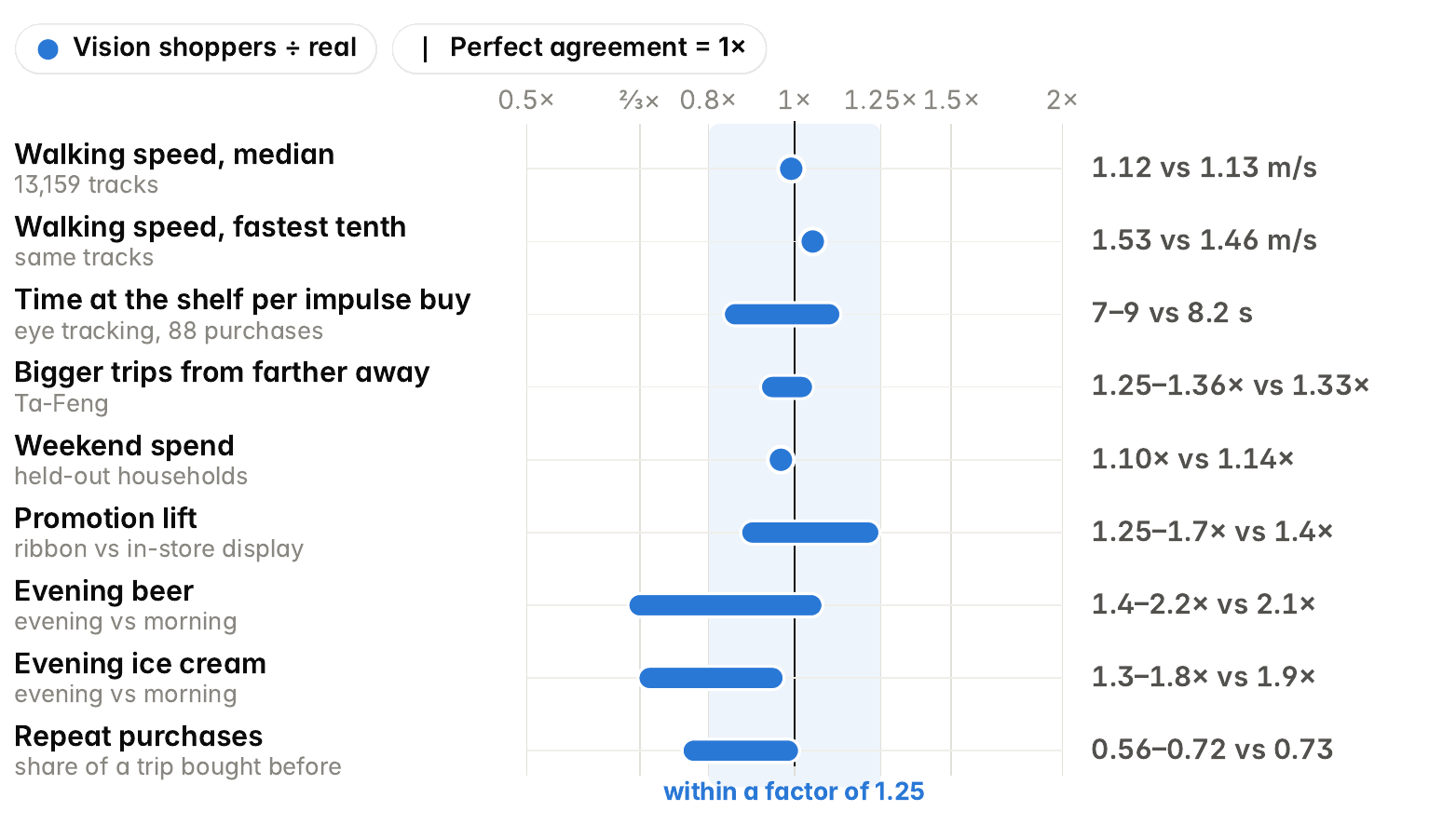}
  \caption{\textbf{Agreement ledger.} Each row divides a simulated behavioral magnitude by its real counterpart; perfect agreement sits on the 1$\times$ line. Bars show the range across decision models and modes.}
  \label{fig:ledger}
\end{figure}

\begin{figure}[!tbp]
  \centering
  \includegraphics[width=\linewidth]{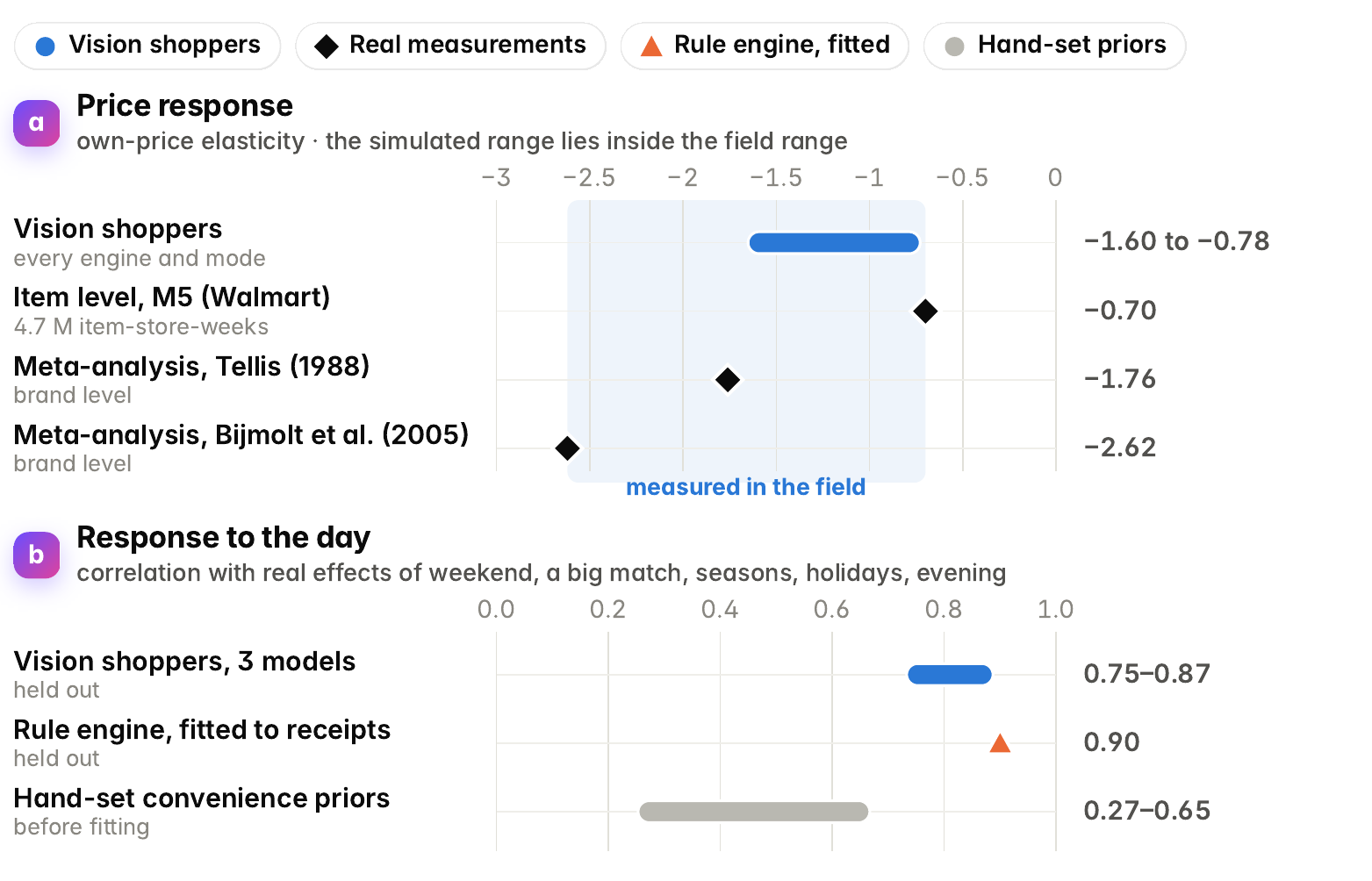}
  \caption{\textbf{Response to the levers.} (a) Simulated own-price elasticities lie inside the range measured in the field. (b) Correlation between simulated and real effects of the day on held-out households.}
  \label{fig:levers}
\end{figure}

\paragraph{Reacting like real shoppers} Simulated own-price elasticities lie between $-0.78$ and $-1.60$ for every engine and mode, inside the field range from item-level sales ($-0.70$) to brand-level meta-analyses ($-1.76$ and $-2.62$;~\citealp{tellis1988,bijmolt2005}). A promotion ribbon lifts its product 1.25--1.7 times, against 1.4 times for an in-store display in the same retailer's data. Customers who live farther away buy 1.25--1.36 times more per trip, against 1.33 in reality; weekend trips are 1.10 times larger, against 1.14 (\cref{fig:ledger}). Effects of weekends, a big match, seasons, holidays and the evening correlate with the real ones at 0.75--0.87 (\cref{fig:levers}).

\paragraph{Moving like real people} On 13,159 pedestrian tracks the median walking speed is 1.13\,m/s; simulated shoppers walk at 1.12\,m/s. Shoppers filmed with eye trackers spent a median 8.2 seconds at the shelf per unplanned purchase; simulated shoppers spend 7--9 seconds.

\paragraph{Replication} Claude Haiku~5.5, Qwen~3.8 Flash and GPT-6 Luna agree on every check with the same personas and settings. A population mixing the three usually lands between the best and the middle model and is the best on product choice and demographic differences.

\section{Implementation and deployment}\label{sec:impl}

\paragraph{Installation} Releases are archives for Windows, macOS and Linux with a single \twin{} icon. The first start sets up the required components and shows its progress; later starts take seconds. No account or key is needed to start, since the built-in sample store runs on the rule-based engine.

\paragraph{Interface} The application works in English, Korean, Chinese, Japanese, Spanish and French, with day and night themes, a command palette that reaches every store and tool, and layouts that keep playback and run controls within reach on phones and tablets.

\paragraph{Models and cost} Before a run, the launch window shows the expected number of model calls. Identical calls can be answered from a response cache. Credentials are stored only on the user's computer, and once a local model has been downloaded the platform runs without a network connection.

\paragraph{Assets} The realistic view uses lifelike avatars released under the MIT License, physically based materials and photogrammetry scans released as CC0, with their sources recorded in the repository.

\paragraph{Quality} Several thousand automated tests cover the engine and the application, including interface checks at phone, tablet and desktop sizes.

\section{Related work}
\paragraph{Synthetic consumers} Language models conditioned on personas reproduce survey response distributions~\citep{argyle2023}, behavioral experiments~\citep{horton2023,aher2023}, market-research estimates~\citep{brand2023,li2024} and experimental outcomes~\citep{hewitt2024}; agents grounded in interviews reproduce individuals' answers~\citep{park2024,toubia2025}, and a causal-inference view cautions about what a prompt changes~\citep{gui2023}. \twin{} moves synthetic consumers into a visual store and validates them on revealed purchases.

\paragraph{Shopping simulation} Agent-based models of grocery shopping~\citep{schenk2007} and path data from instrumented stores~\citep{hui2009path,hui2009jcr} established routes, stops and baskets as the structure of a trip; generative agents~\citep{park2023} added language-model reasoning in simulated spaces. In-store research shows that facings and position drive attention and choice~\citep{chandon2009}, and vision--language models~\citep{alayrac2022,liu2023} make it possible for synthetic shoppers to respond to them. Digital twins~\citep{grieves2017} provide the frame for trying decisions on a mirror first.

\section{Limitations}
The size of a household's trip is taken from its receipts, so interventions in the current validation move purchases between products within a trip, and extending the validation to effects that work through additional store visits is a natural next step. The real price responses come from other retailers and from meta-analyses, and promotion lifts are observational. The main test covers one store format, a US supermarket, and convenience-store situation effects rest on priors until public convenience-store receipts become available. Synthetic consumers inherit the knowledge and biases of the models that drive them, so \twin{} is designed to screen alternatives and sharpen hypotheses, narrowing the options that go on to a test with real customers.

\section{Availability and license}
\twin{} is available at \url{https://github.com/BrainJellyPie/twinstore} with installers for Windows, macOS and Linux. Its source code is published under the \twin{} License~1.0, a source-available license that permits research and non-commercial use; the full terms are in the repository. \twin{} builds on Parastore, whose MIT License notice is retained.

\section{Conclusion}
\twin{} brings synthetic consumers into a model of the physical store and lets them decide from what they see. Stores are built from formats, floor plans or photographs; shoppers who look at the shelves produce receipts for configurations that have not been deployed; and the platform's tools turn those receipts into paired comparisons, layout searches, interviews and studies. On held-out purchase data the shoppers approach the households' own natural variation and respond to price, promotion, distance and the day with magnitudes close to those measured in the field, across three independent models.

\paragraph{Acknowledgments} We thank intellicia for releasing Parastore and the providers of the public datasets used in the validation.

\clearpage
\normalsize
\appendix

\section{Reading the screens}\label{app:intro}

The main text describes \twin{} by its components and tests its output against real purchases. This appendix shows the platform at work, screen by screen, and makes an argument that complements the validation. Every link in the chain from a real store to a simulated receipt is visible in the interface and can be checked by the person who uses it. \Cref{sec:validation} establishes how close the receipts come to real ones; the screens below show how the receipts are produced, so that a reader can see why they come out as they do.

\paragraph{Provenance} All screens were captured from the public build of October 2026 with the sample store that ships with the platform: a convenience store open on weekdays from 8\,am to 8\,pm, with 2,440 synthetic customers over five weekdays and two layouts. The sample store carries two stored runs of the language engine, one per layout, on the same customers; for this appendix we also ran the rule-based engine on all 2,440 customers. The images that a vision shopper receives (\cref{app:see}) were produced by the engine's own renderer from the same store, and the Shelf~Lab edit (\cref{app:capture}) by the platform's edit engine on the store's stored shelf photograph. Apart from cropping and the numbered markers, no screen was edited, and every number quoted below can be read off a screen or recomputed from a run's per-shopper export.

\paragraph{The argument} A receipt from a synthetic shopper is informative about a store that has not changed only if five conditions hold.
\begin{enumerate}[label=\textbf{C\arabic*}]
  \item \textbf{The twin is the store in question.} Its layout, assortment, prices, signs, hours and neighborhood are those of the store the decision concerns.
  \item \textbf{The shopper receives what a person in the aisle would see.} A change to a pack, a tag or a sign reaches the decision as a change in its input, and nothing else changes with it.
  \item \textbf{Every decision can be traced.} Each purchase is tied to a time, a shelf, a product and a stated reason, so that an implausible trip can be found.
  \item \textbf{The output is in the retailer's own terms.} Trips aggregate to the receipts, revenue, conversion and shelf sales a retailer already reads, and can be held against the retailer's own records.
  \item \textbf{A difference between alternatives is caused by the change.} The same shoppers meet both alternatives, and the uncertainty of the difference is reported with it.
\end{enumerate}
\Cref{tab:appx-map} lists, for each condition, the screens that make it visible and a check that anyone can repeat on the sample store. \Cref{sec:validation} measures how close the resulting receipts come to real ones, and the screens add that a user does not have to take the mechanism on trust.

\begin{table}[!htbp]
\centering\small
\caption{\textbf{Conditions for an informative synthetic receipt}, where each is visible, and a check on the sample store.}
\label{tab:appx-map}
\begin{tabularx}{\linewidth}{@{}>{\sffamily\bfseries}l>{\raggedright\arraybackslash}p{0.17\linewidth}>{\raggedright\arraybackslash}X@{}}
\toprule
 & \sffamily Screens & \sffamily A check on the sample store\\
\midrule
C1 & \cref{fig:appx-build,fig:appx-capture,fig:appx-town} & Every fixture, product, price and opening hour is a record that can be read, edited and exported; the 3D view renders the same records the shoppers use.\\
C2 & \cref{fig:appx-see,fig:appx-capture} & A promotion added to a rendered shelf face changes 0.8\,\% of its pixels, all in the ribbon and the tag; an edit of a shelf photograph changes no pixel outside the edited regions.\\
C3 & \cref{fig:appx-journey} & Any of the 2,440 shoppers can be opened; each purchase shows its time, shelf, price and the reason recorded at that step.\\
C4 & \cref{fig:appx-day,fig:appx-results} & The revenue on screen equals the sum over the per-shopper export (\$21,032.06 over 2,440 rows); receipts, POS files and tracks can be uploaded and compared with the simulation.\\
C5 & \cref{fig:appx-compare,fig:appx-ab} & The layout comparison pairs 2,440 shoppers and reports intervals; unpaired, the same difference would not be distinguishable from noise.\\
\bottomrule
\end{tabularx}
\end{table}

\section{Building the twin}\label{app:build}

\begin{figure}[!tbp]
  \centering
  \includegraphics[width=\linewidth,height=0.74\textheight,keepaspectratio]{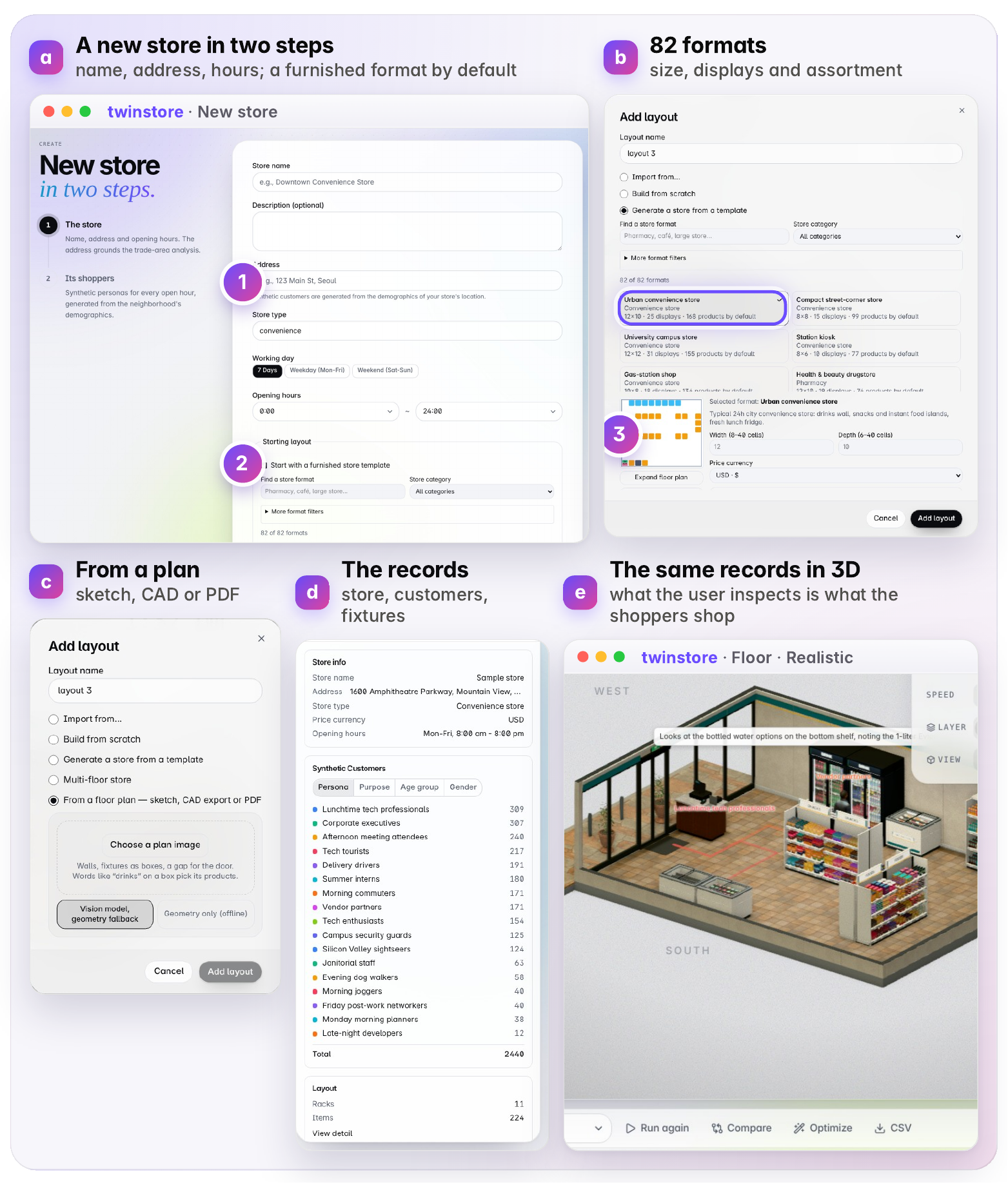}
  \caption{\textbf{Building the twin.} (a) The new-store page: name, address, type, working days and hours; the address grounds the trade area (1), and a furnished format is the default starting layout (2). (b) The 82 formats, each with its grid size, displays and default assortment, and a preview of the floor plan (3). (c) A store from a plan image, read by a vision model with a geometric fallback. (d) The store's records: information, synthetic customers by persona, fixtures and products. (e) The same records rendered in 3D.}
  \label{fig:appx-build}
\end{figure}

\paragraph{A store from a format} The new-store page (\cref{fig:appx-build}a) asks for what defines a store to its customers: a name, an address, a type, the working days and the opening hours. The address is not decorative. It grounds the trade-area model of \cref{app:town}, and in the second step the synthetic customers of every open hour are generated from the neighborhood's demographics (marker~1). By default the floor is furnished from one of the 82 formats (marker~2). The format list (\cref{fig:appx-build}b) states what each format brings---the urban convenience store, for example, is a $12\times10$ grid with 25 displays and 168 products by default; the station kiosk is $8\times6$ with 10 displays and 77 products---and a preview of the floor plan, with its width, depth and price currency, appears before anything is created (marker~3).

\paragraph{A store from a plan} When the store already exists on paper, a plan image---a hand sketch, a CAD export or a PDF page---is read into the grid (\cref{fig:appx-build}c). The instructions on the screen are the whole convention: walls, fixtures drawn as boxes, a gap for the door, and words such as ``drinks'' on a box to choose its products. A vision model reads the image, with a geometric reader as a fallback; the geometric reader alone works offline.

\paragraph{Records, not pictures} The Store tab (\cref{fig:appx-build}d) lists what the store consists of: its name, address, type, currency and opening hours (Monday to Friday, 8\,am to 8\,pm in the sample store); the composition of its 2,440 synthetic customers by persona, purpose, age and gender, from 309 lunchtime tech professionals and 307 corporate executives down to 12 late-night developers; and the 11 fixtures and 224 products of the layout. The 3D view (\cref{fig:appx-build}e) is a rendering of these records, not a separate model. This matters for C1. When a scene for people and a parameter file for the simulator are kept separately, the two can drift apart unnoticed; in \twin{} the store a manager inspects is, by construction, the store the shoppers shop.

\section{From a shelf photograph to a planogram and its counterfactuals}\label{app:capture}

\begin{figure}[!tbp]
  \centering
  \includegraphics[width=\linewidth,height=0.74\textheight,keepaspectratio]{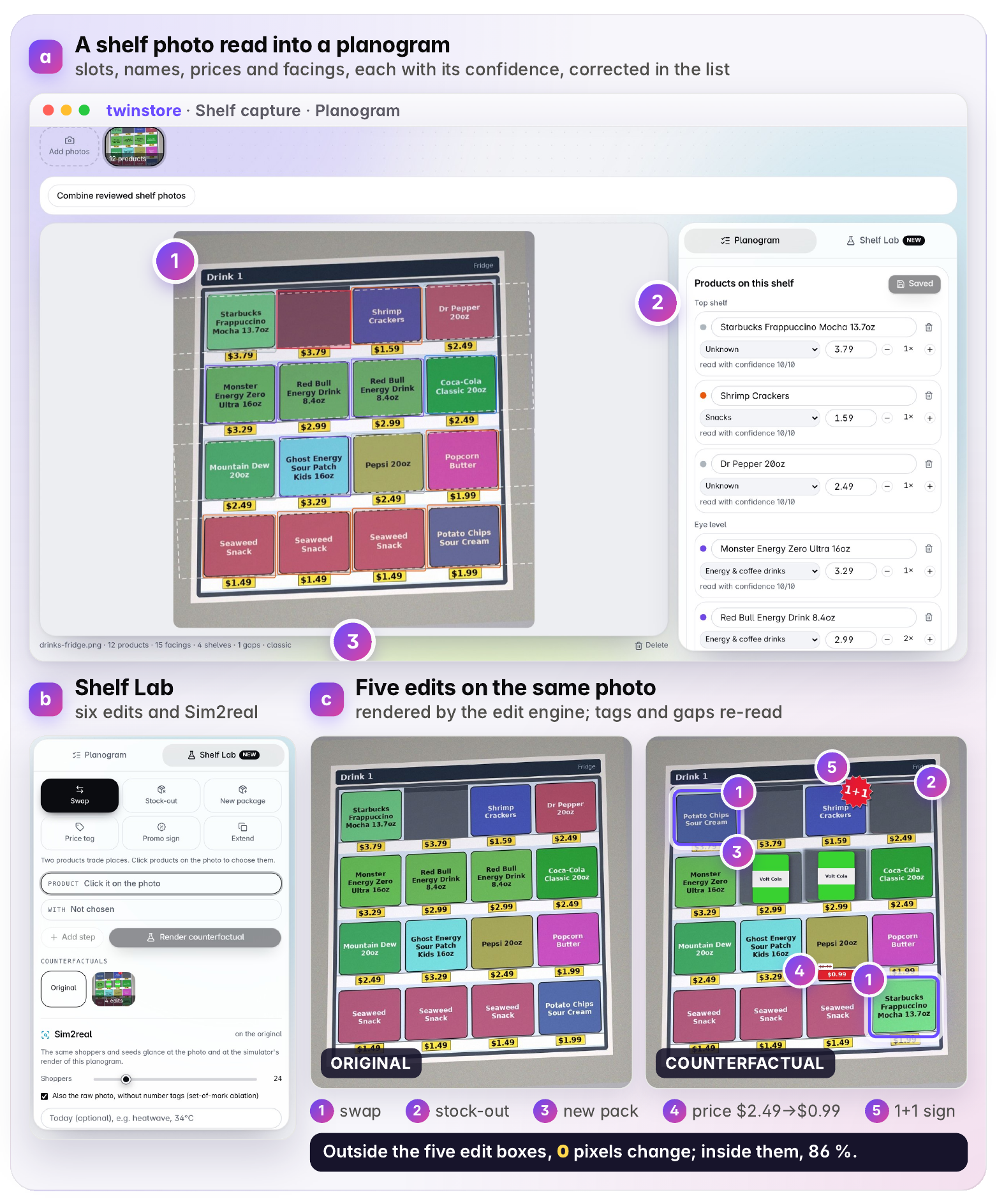}
  \caption{\textbf{From a shelf photograph to a planogram and its counterfactuals.} (a) The sample store's fridge photograph after reading: detected slots on the photo (1), each product with its category, price, facings and reading confidence (2), and the summary of 12 products, 15 facings, 4 shelves and 1 gap (3). (b) Shelf Lab's six edits and the Sim2real comparison. (c) Five edits applied by the edit engine: a swap (1), a stock-out (2), a new pack on two facings (3), a price cut shown against the old price (4) and a 1+1 sign (5). No pixel outside the edited regions changes.}
  \label{fig:appx-capture}
\end{figure}

\paragraph{Reading the photograph} \Cref{fig:appx-capture}a shows the sample store's fridge photograph after reading. Every detected slot is outlined on the photo (marker~1), and the list beside it gives, for every product, its name, category, price and number of facings, with the confidence of the reading (marker~2); the empty slot is marked as a gap. The summary line records the outcome: 12 products, 15 facings, 4 shelves and 1 gap (marker~3). Every entry can be corrected before it is used, and the reviewed shelf can then be added as a new fixture, replace an existing one, start a new layout or be compared with the store's planogram. Review before use keeps every simulation input under the user's control, and a reviewed reading makes the twin's prices and facings those of the store's own shelf instead of a template's, which is condition C1 at the level of a single fixture.

\paragraph{Changing the photograph} Shelf Lab (\cref{fig:appx-capture}b) offers six edits on the same photograph: swap two products, empty a slot, paste a new package, change a price tag, add a promotion sign and extend a row. \Cref{fig:appx-capture}c applies five of them at once. Two products trade places (1); Dr Pepper is removed, leaving a gap (2); a new pack, ``Volt Cola'', replaces Red Bull on both facings (3); Pepsi's tag drops from \$2.49 to \$0.99 with the old price struck through (4); and a ``1+1'' burst is placed on the shrimp crackers (5). The edit is local by construction: outside the five edited regions no pixel changes, and inside them 86\,\% do. The edited photograph is read again, so its tags and gaps enter the simulation like those of any other capture.

Sim2real, at the bottom of panel~(b), sends the same shoppers with the same seeds to the photograph, to the simulator's render of the same planogram and, optionally, to the raw photograph without number tags. The effect of rendering on decisions is thereby measured rather than assumed. For condition C2 the conclusion is simple: a test of a visual change is only as clean as the change, and when the counterfactual differs from the original in the edited pixels alone, any difference in the shoppers' choices can be attributed to the edit.

\section{The simulated day}\label{app:day}

\begin{figure}[!tbp]
  \centering
  \includegraphics[width=\linewidth]{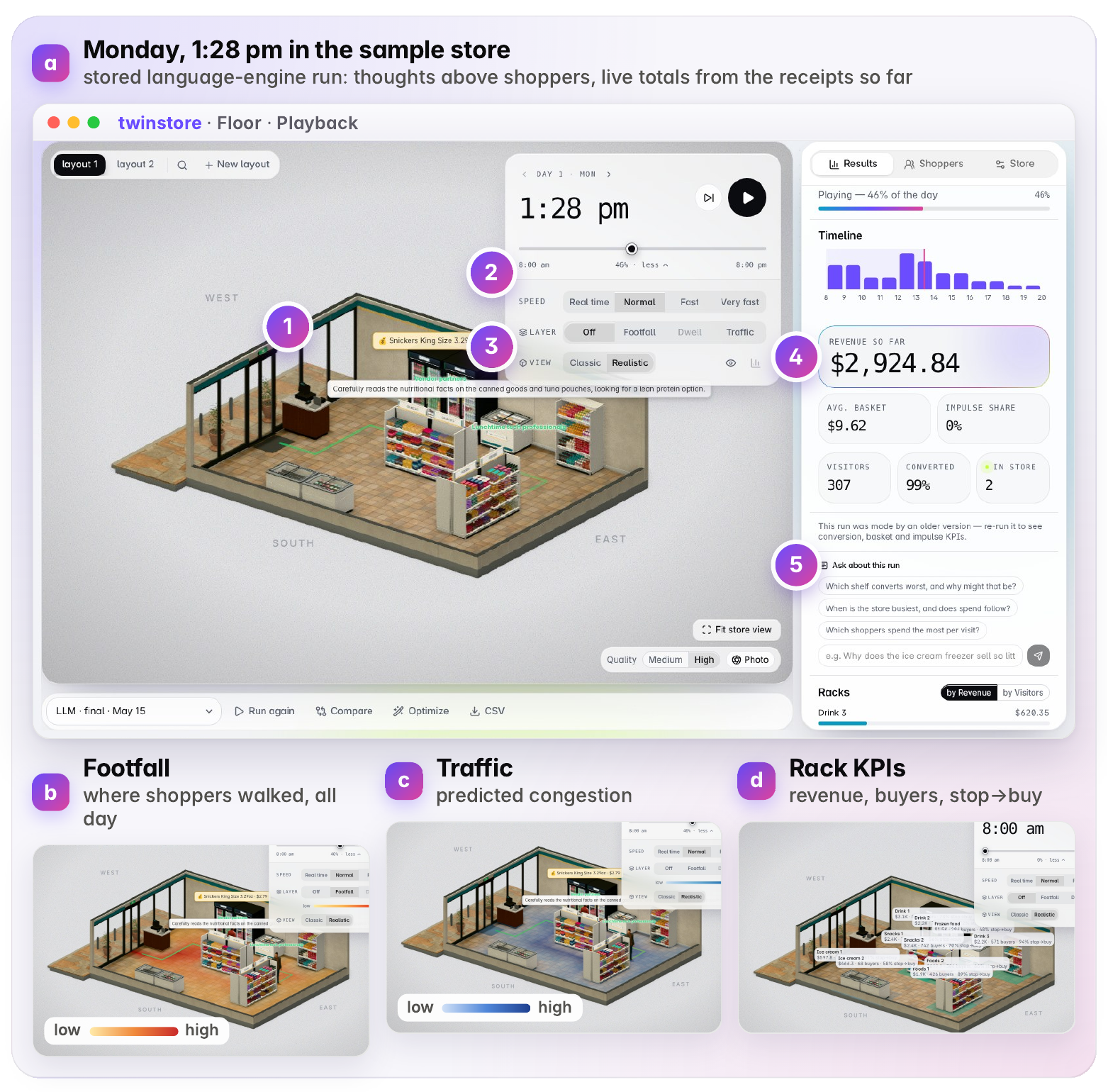}
  \caption{\textbf{The simulated day.} (a) Monday at 1:28\,pm in the stored language-engine run of the sample store: a recorded thought and a pick above the shoppers (1), the clock and speed (2), layers and views (3), live totals of the receipts so far (4) and questions about the run (5). (b) Footfall and (c) predicted congestion over the day. (d) Revenue, buyers and stop-to-buy rate per fixture, from the rule-based run on all 2,440 shoppers.}
  \label{fig:appx-day}
\end{figure}

\paragraph{Playback} \Cref{fig:appx-day}a is the sample store on Monday at 1:28\,pm, replayed from the stored language-engine run. Above a shopper at the canned goods is the thought recorded at that step---``Carefully reads the nutritional facts on the canned goods and tuna pouches, looking for a lean protein option''---and a label shows a pick as it happens, a Snickers King Size at \$2.79 (marker~1). The clock, day and speed (2) and the layer and view switches (3) control the replay. The results panel follows it (4): by 1:28\,pm, 307 visitors have produced \$2,924.84 in revenue with an average basket of \$9.62, and two shoppers are in the store. Questions about the run can be typed beneath (5); the answers are drawn from the run's own fact sheet and list the figures they rest on.

\paragraph{Layers} The same day can be read in aggregate. Footfall (\cref{fig:appx-day}b) shows where shoppers walked and traffic (c) where congestion is predicted. Rack KPIs (d), from the rule-based run on all 2,440 shoppers, label each fixture with its revenue, buyers and stop-to-buy rate: the drinks fridge \emph{Drink~3} converts 94\,\% of the shoppers who stop at it and 571 of them buy; the second snack shelf takes \$2.4K from 742 buyers at 70\,\%; the frozen-food case converts 48\,\% and the second ice-cream freezer 58\,\% of 68 buyers. These are quantities a store manager already reasons about; the simulation adds that they can be compared before and after a change.

A person who knows the store can judge plausibility at a glance: where the crowds form, which shelves are passed by, when the store is busy. The footfall map is also the quantity that \cref{sec:validation} compares with tracked pedestrians and that a user can compare with their own tracks (\cref{app:results}).

\section{Configuring a run}\label{app:run}

\begin{figure}[!tbp]
  \centering
  \includegraphics[width=\linewidth]{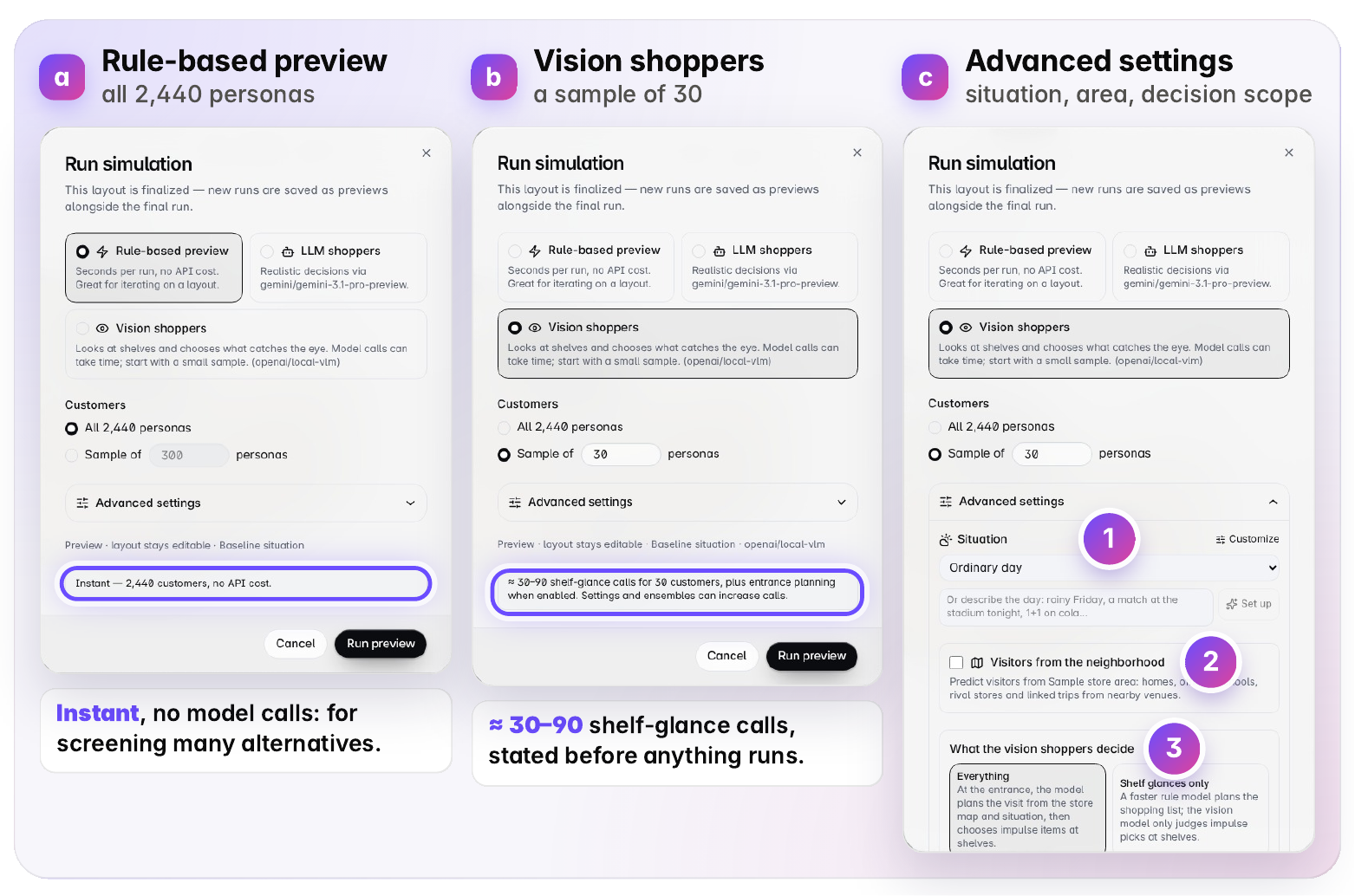}
  \caption{\textbf{Configuring a run.} (a) The rule-based preview on all 2,440 personas, with no model calls. (b) Vision shoppers on a sample of 30, with the expected number of calls stated before the run. (c) Advanced settings: the situation of the day, preset or described in words (1), visitors predicted from the neighborhood (2), and what the vision shoppers decide (3).}
  \label{fig:appx-run}
\end{figure}

The run window states three things before anything runs: the engine, the population and the cost. The rule-based preview on all 2,440 personas is instant and makes no model calls (\cref{fig:appx-run}a); it is the engine for screening many alternatives. Vision shoppers on a sample of 30 are announced as about 30--90 shelf-glance calls, plus entrance planning when enabled (b); for the language engine on 300 customers the window states about 900 calls, three per customer, and notes that identical prompts can be answered from the response cache. Advanced settings (c) set the situation of the day---a preset such as an ordinary day, or a description in words such as ``rainy Friday, a match at the stadium tonight, 1+1 on cola'' (1)---whether visitors are predicted from the neighborhood's homes, offices, schools, rival stores and linked trips (2), and what the vision shoppers decide (3): everything, from the entrance plan to the impulse picks, or only the shelf glances, with the shopping list planned by the faster rule model. Further down are the model---the server default, one model or a weighted ensemble---the seed and replicates.

The seed and the response cache make a run a function of its inputs. Re-running a configuration reproduces it; a replicate with a new seed measures run-to-run noise; and two runs with the same seed share every random draw, which is what makes the comparisons of \cref{app:compare} paired. Stating the number of calls in advance lets the user choose the cheapest engine that can answer the question at hand, and keeps the cost of a study visible to whoever approves it.

\section{What a vision shopper sees}\label{app:see}

\begin{figure}[!tbp]
  \centering
  \includegraphics[width=\linewidth,height=0.74\textheight,keepaspectratio]{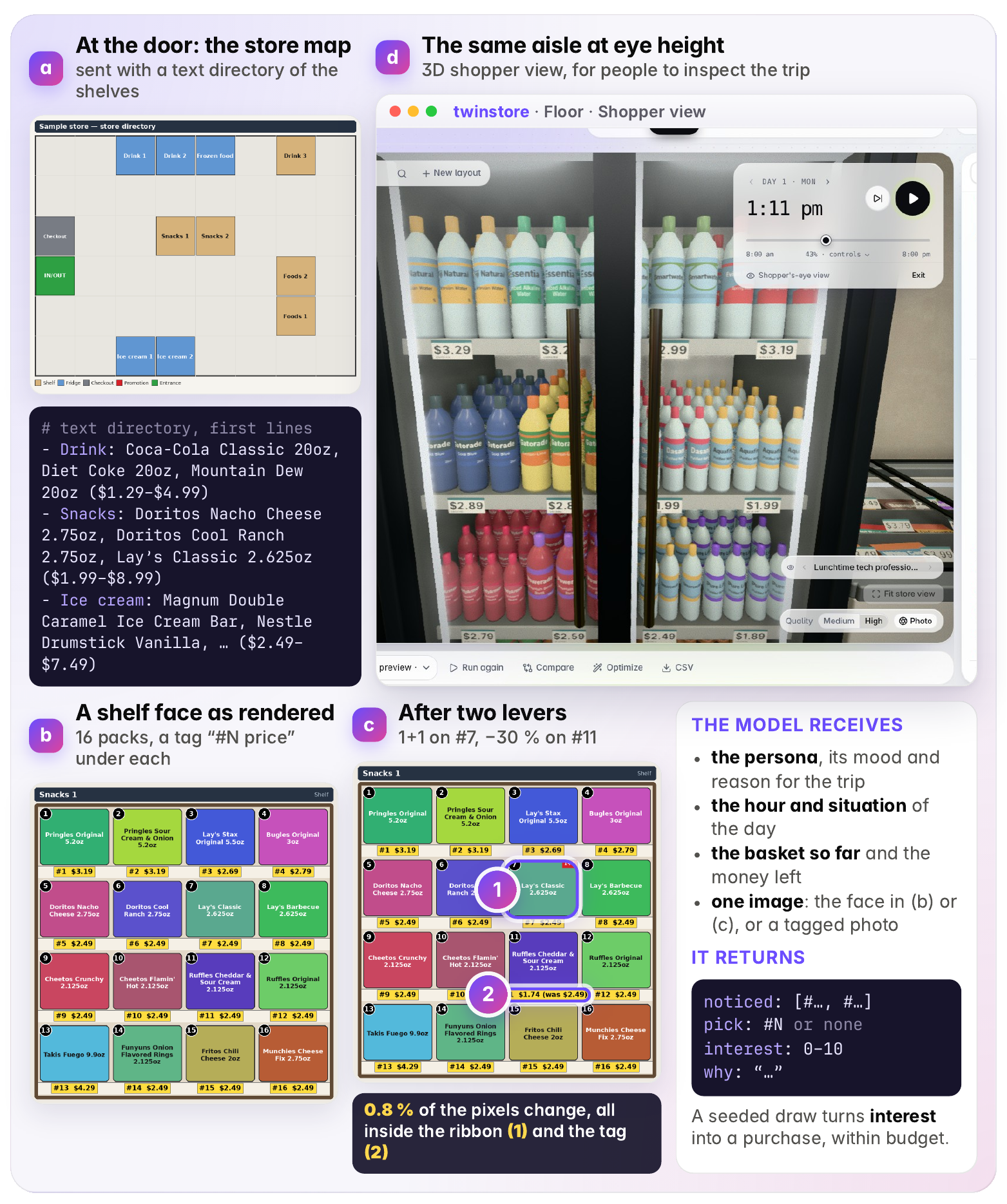}
  \caption{\textbf{What a vision shopper sees}, as rendered by the engine from the sample store. (a) The store map and text directory received at the door. (b) A shelf face with a tag ``\#$N$ price'' under every pack. (c) The same face with a 1+1 ribbon on \#7 (1) and a 30\,\% cut on \#11 (2): 0.8\,\% of the pixels change. (d) The 3D shopper's-eye view, used to inspect trips. Bottom right: what the model receives with each image and the structure of its answer.}
  \label{fig:appx-see}
\end{figure}

\Cref{fig:appx-see} shows the inputs of the vision engine as the model receives them. At the door (a), the shopper receives a schematic map of the store with each fixture named by its category, together with a text directory that gives three example products and the price range of each category. Along the route, the twin renders each well-seen shelf face (b): sixteen packs with their printed names, a badge number on each pack and, beneath it, a yellow tag ``\#$N$ price'' that ties number, name and price together on the line the model reads. The model receives this image with the persona, its mood and reason for the trip, the hour and the situation, the basket so far and the money left, and returns the products it noticed, the one it would pick, an interest rating from 0 to 10 and a reason. A seeded draw turns the rating into a purchase within the budget.

Panel (c) is the same face after two levers: a ``1+1'' ribbon on Lay's Classic (\#7) and a 30\,\% cut on Ruffles Cheddar \& Sour Cream (\#11), shown as ``\$1.74 (was \$2.49)''. Only 0.8\,\% of the pixels change, all inside the ribbon and the tag. Two consequences follow. First, the lever reaches the decision where a person would meet it, on the shelf, rather than as a word in a product list; this is condition C2 and design goal G1. Second, because model responses are cached by their full input, image included, a shopper who never glances at this face receives exactly the same inputs in both arms and makes exactly the same decisions. Differences between the arms can arise only from glances at the changed face and from what follows them.

The 3D shopper's-eye view (d) shows the store from a shopper's eye height, with packs and price tags on the shelves. It serves people who inspect a trip; the model works from the shelf renders in (b) and (c) or from a tagged photograph.

\section{Following one shopper}\label{app:journey}

\begin{figure}[!tbp]
  \centering
  \includegraphics[width=\linewidth]{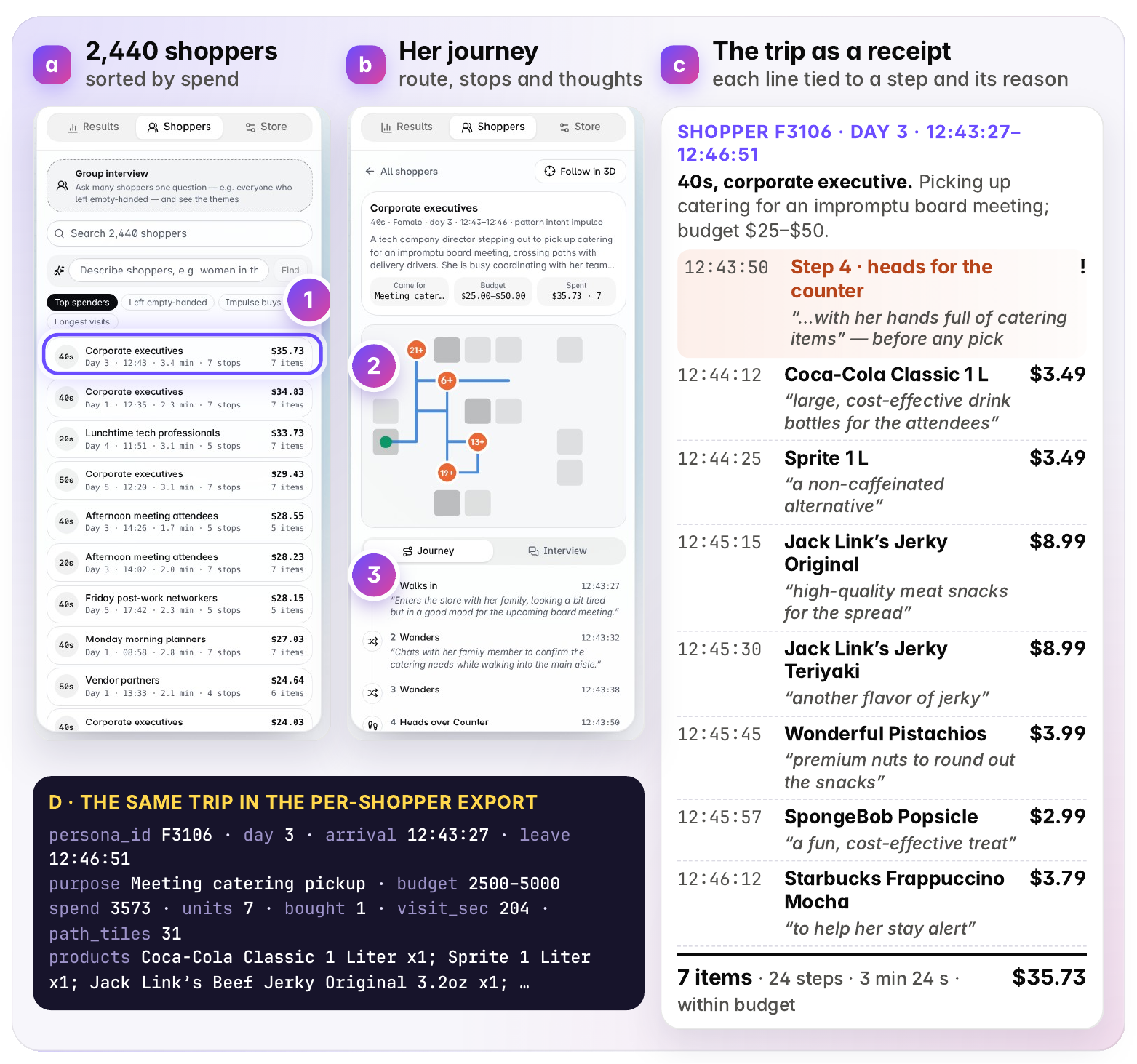}
  \caption{\textbf{Following one shopper} in the stored language-engine run. (a) The 2,440 shoppers sorted by spend; the top shopper (1). (b) Her persona, route and stops (2) and the steps with the thoughts recorded at them (3). (c) The trip as a receipt, each line tied to its time and reason; the marked step~4 shows a narrated step checked against the picks. (d) The same trip as a row of the per-shopper export.}
  \label{fig:appx-journey}
\end{figure}

The Shoppers tab (\cref{fig:appx-journey}a) lists all 2,440 shoppers of a run and sorts them by spend, empty-handed exits, impulse purchases or time in store; a sentence such as ``women in their thirties who left empty-handed'' selects a group, and a group interview puts one question to many of them. Sorted by spend, the top shopper of the five days is a corporate executive in her forties on day~3 at 12:43 (marker~1). Her journey (b) opens with the persona as generated---``a tech company director stepping out to pick up catering for an impromptu board meeting'', with a budget of \$25--\$50---draws her route and stops on the floor plan (2), and lists every step with its time and the thought recorded at it (3).

Panel (c) arranges the steps as the receipt they produce. At the drinks fridge she takes a one-liter Coca-Cola, then ``looks for a non-caffeinated alternative'' and takes a Sprite; at the snack shelf, two flavors of jerky and pistachios ``to round out the snacks''; a popsicle at the freezer; and a Frappuccino for herself ``to help her stay alert''. Seven items, \$35.73, within budget, in 3 minutes 24 seconds. The same trip is one row of the per-shopper export (d), the table from which every aggregate on screen and every number in the validation are computed.

Because every step is recorded with its time and thought, each narrated step can be checked against the trip. At step~4 the narration has her approach the counter ``with her hands full of catering items'' before her first pick; a receipt log would not show such a mismatch, while the journey makes it visible at once. This is condition C3 at work, and the validation in \cref{sec:validation} accordingly scores the receipts themselves. The tab next to the journey interviews the shopper about the trip; answers cite the steps they rely on, and those steps are highlighted on screen.

\section{From trips to results}\label{app:results}

\begin{figure}[!tbp]
  \centering
  \includegraphics[width=\linewidth]{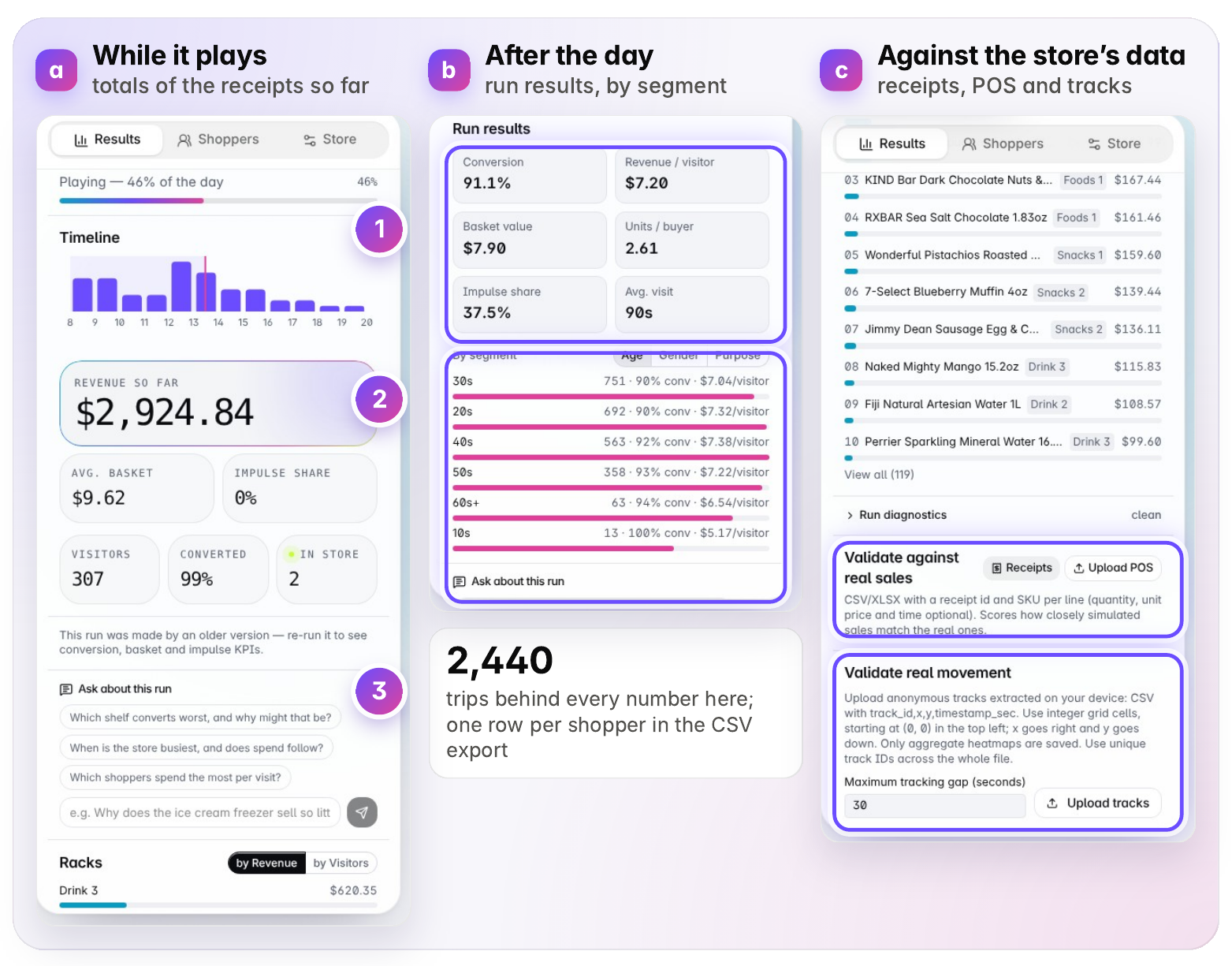}
  \caption{\textbf{From trips to results.} (a) While a day plays: visitors by hour (1), totals of the receipts so far (2) and questions about the run (3). (b) Run results of the rule-based run on all 2,440 shoppers, with a breakdown by age. (c) Products ranked by revenue, run diagnostics, and the upload of the store's receipts, POS exports and movement tracks for comparison with the simulation.}
  \label{fig:appx-results}
\end{figure}

Every number on the results panels is an aggregate of trips. While a day plays (\cref{fig:appx-results}a), the timeline of visitors by hour (1), the revenue, basket, visitors and conversion so far (2) and the questions about the run (3) refer to the receipts produced up to the current minute. After the day, the results of the rule-based run on all 2,440 shoppers (b) give conversion (91.1\,\%), revenue per visitor (\$7.20), basket value (\$7.90), units per buyer (2.61), impulse share (37.5\,\%) and average visit (90\,s), and break them down by age, gender or purpose: shoppers in their twenties and thirties convert at 90\,\%, those in their fifties at 93\,\%, and revenue per visitor lies between \$7.04 and \$7.38 in the four largest age groups. The ranking of products by revenue, buyers or quantity follows (c), with the run diagnostics, which check for unprocessed shoppers and trips over budget and report this run as clean.

The lower part of (c) closes the loop with the store's own data. Receipts or point-of-sale exports, one line per receipt and product, are uploaded and scored against the simulated sales; anonymous movement tracks, extracted on the user's device, are compared with the simulated footfall, and only aggregate heatmaps are kept. These are the comparisons that \cref{sec:validation} makes on public data, available to every user on their own store. This is condition C4: because the simulation speaks in the units of the retailer's records, it can be held to them. In the sample store the check is exact at the level of totals: the revenue the panel reports for the stored run, \$21,032.06, is the sum of the spend column over the 2,440 rows of its export.

\section{Comparing alternatives on the same shoppers}\label{app:compare}

\begin{figure}[!tbp]
  \centering
  \includegraphics[width=\linewidth,height=0.74\textheight,keepaspectratio]{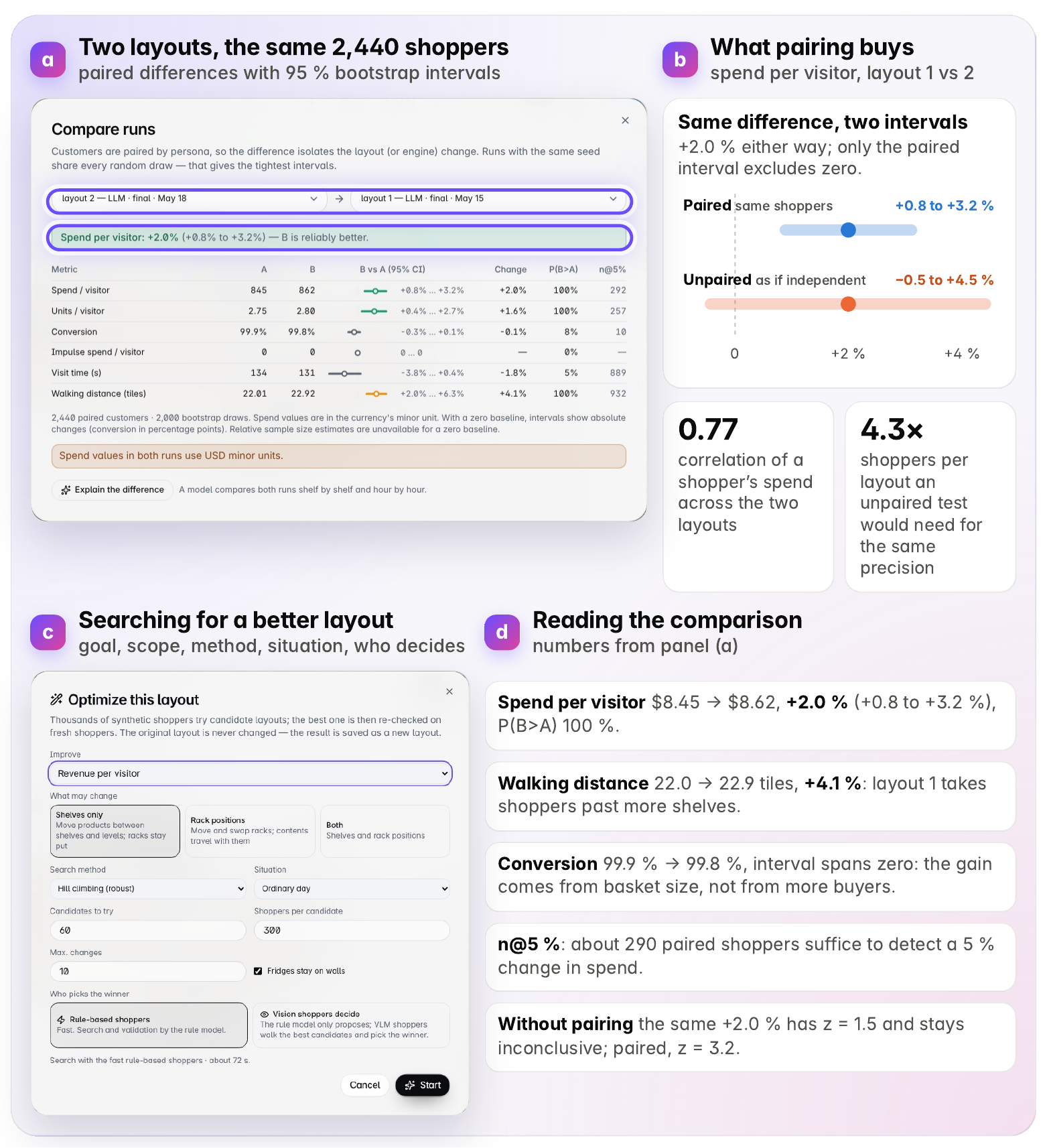}
  \caption{\textbf{Comparing two layouts on the same shoppers.} (a) The comparison window for the two stored runs of the sample store: the selected runs and the verdict (rings), then every metric with both means, the paired difference and its 95\,\% interval, the probability that B is better and the paired sample a 5\,\% effect would need. (b) The spend difference with the paired interval and with the interval two independent samples of the same size would give, computed from the runs' per-shopper exports. (c) The layout optimizer. (d) The comparison read row by row.}
  \label{fig:appx-compare}
\end{figure}

The two layouts of the sample store were run with the same 2,440 personas, and \cref{fig:appx-compare}a compares them. Each shopper's trip in one layout is paired with the same shopper's trip in the other, and the window reports, for every metric, both means, the paired difference with a 95\,\% bootstrap interval, the probability that B is better and the number of paired shoppers a 5\,\% effect would need. The verdict is in the first line: spend per visitor is 2.0\,\% higher in layout~1 (95\,\% interval $+0.8$ to $+3.2$\,\%), \$8.62 against \$8.45. The other rows say where the difference comes from (d). Conversion does not change (99.8\,\% against 99.9\,\%, an interval that spans zero); units per visitor rise by 1.6\,\%; shoppers walk 4.1\,\% farther. Layout~1 takes shoppers past more shelves and they buy slightly more; it does not bring in more buyers. A manager reading this window learns not only which layout is better but by which mechanism, and therefore what to watch if it is built.

Panel (b) shows what pairing contributes. A shopper's spend in the two layouts is correlated at 0.77, because the same person with the same random draws makes mostly the same decisions. The paired interval is therefore about half as wide as that of two independent samples of the same size: the standard error of the difference is 5.2 cents against 10.9. Unpaired, the 2.0\,\% difference would have $z=1.5$ and an interval from $-0.5$ to $+4.5$\,\%, indistinguishable from noise; paired, $z=3.2$. Reaching the same precision without pairing would take about 4.3 times as many shoppers per layout. This is condition C5 and design goal G2 in numbers: the comparison isolates the change, and it needs fewer shoppers to do so.

The optimizer (c) turns comparisons into a search. The user chooses the goal---revenue per visitor, conversion, units, impulse revenue or shorter paths---what may change (products between shelves and levels, fixture positions, or both), the search method, the situation, the number of candidates (60), shoppers per candidate (300) and the largest number of changes (10), and can keep fridges on walls. Rule-based shoppers search, in about 72 seconds in this configuration, and either they or vision shoppers pick the winner on fresh shoppers. The original layout is never changed; the best candidate is saved as a new layout, which can then be compared with the original like any other.

\section{The marketing studio}\label{app:studio}

\begin{figure}[!tbp]
  \centering
  \includegraphics[width=0.9\linewidth,height=0.76\textheight,keepaspectratio]{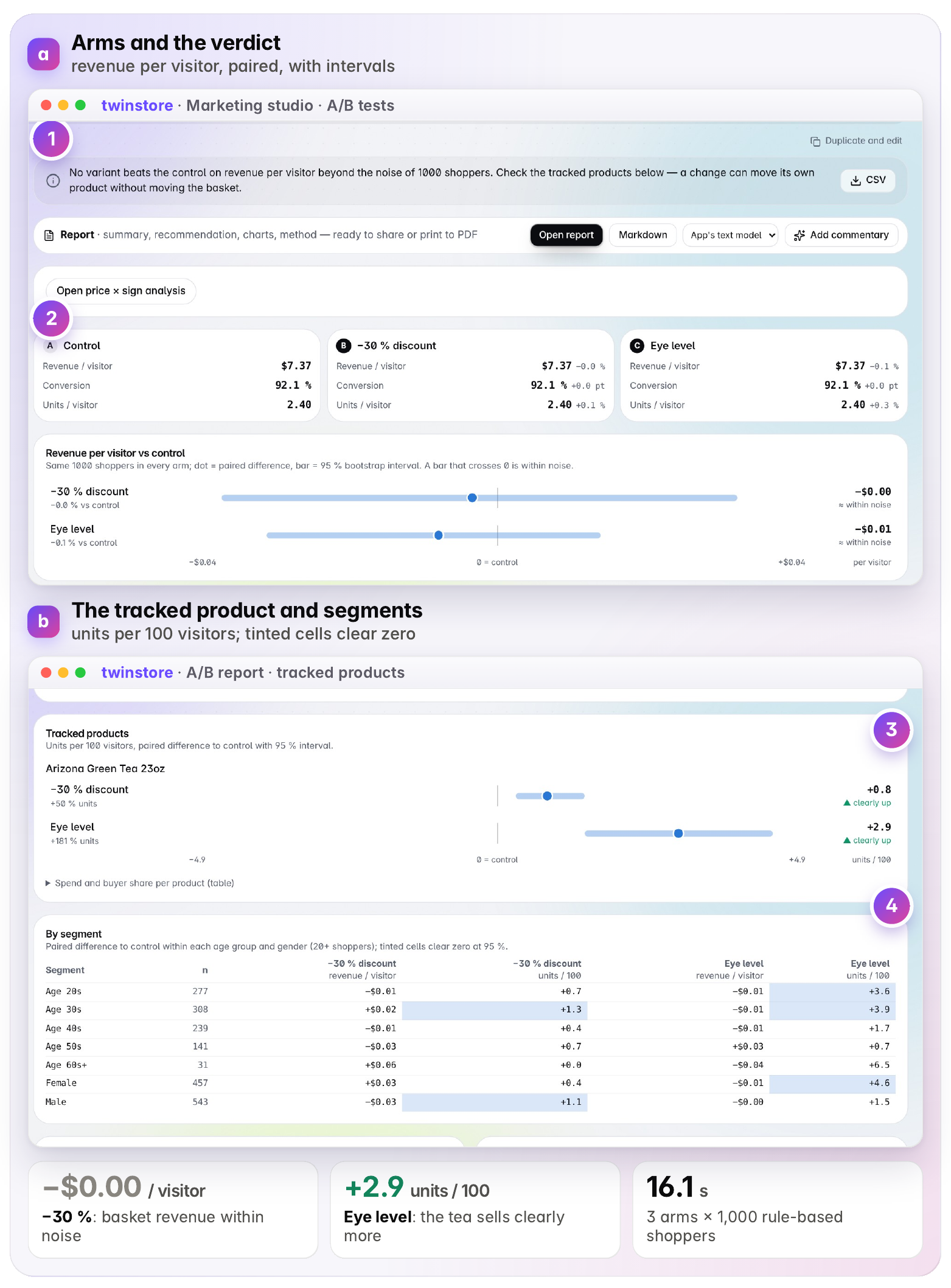}
  \caption{\textbf{An A/B test from the marketing studio}: a 30\,\% discount and a move to eye level for one tea, each against the unchanged layout with the same 1,000 rule-based shoppers. (a) The verdict (1) and the arms with revenue per visitor and its paired intervals (2). (b) Units of the tested product per 100 visitors (3) and the response by segment (4).}
  \label{fig:appx-ab}
\end{figure}

\paragraph{An A/B test and its report} \Cref{fig:appx-ab} is a stored A/B test of the sample store: a 30\,\% discount on Arizona Green Tea (B) and a move of the same tea to eye level (C), each against the unchanged layout (A), with the same 1,000 rule-based shoppers in every arm. It ran in 16.1 seconds. The verdict (1) finds that no variant beats the control on revenue per visitor beyond the noise of 1,000 shoppers. All three arms take \$7.37 per visitor, and the intervals of both differences cross zero (2). The report does not stop there. The tracked product moves clearly (3): the discount sells 0.8 more teas per 100 visitors (+50\,\%), eye level 2.9 more (+181\,\%), both with intervals above zero. The segment table (4) locates the response: eye level works for shoppers in their twenties and thirties and for women, the discount for shoppers in their thirties and for men.

The first line of the report states the managerially relevant reading: a change can move its own product without moving the basket. A tool that reported only the product's lift would have recommended both changes; a tool that reported only revenue per visitor would have missed that placement moved the tea at all. Reporting both, with their uncertainty, on the same shoppers, is what makes the result usable. Every study can be duplicated with new conditions, exported, and turned into a report with charts, method and an optional written commentary.

\begin{figure}[!tbp]
  \centering
  \includegraphics[width=\linewidth,height=0.74\textheight,keepaspectratio]{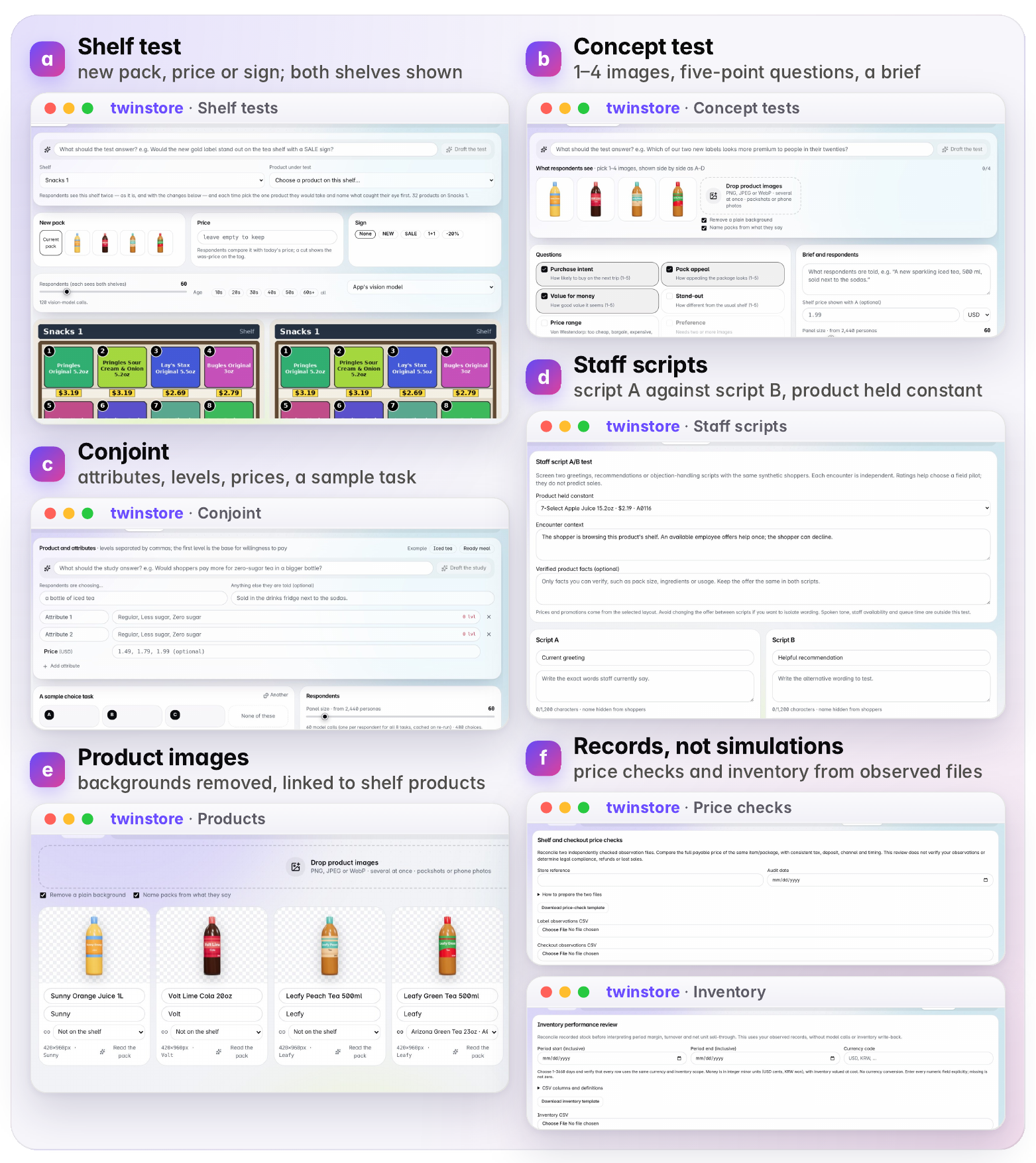}
  \caption{\textbf{The other study builders}: (a) shelf test, (b) concept test, (c) conjoint, (d) staff scripts, (e) product images, and (f) price checks and inventory reviews, which work on observed records without model calls.}
  \label{fig:appx-studio}
\end{figure}

\paragraph{The other tools} \Cref{fig:appx-studio} shows the builders of the remaining tools, each at its set-up stage. A shelf test (a) shows respondents the same shelf twice, as it is and with a new pack, price or sign, and asks each time which product they would take and what caught their eye first; 60 respondents mean 120 vision-model calls, stated before the test runs. A concept test (b) shows one to four pack images side by side and asks purchase intent, pack appeal, value for money, stand-out and a Van Westendorp price range. A conjoint study (c) takes attributes, levels and prices, shows a sample choice task, and states its cost: 60 model calls for 60 respondents and 480 choices, cached on re-run. A staff-script test (d) compares two scripts on the same shoppers with the product held constant, and its ratings guide which script goes on to a field pilot. Product images (e) have their backgrounds removed and are linked to products on the shelves, where they appear on the packs in 3D and in the shoppers' renders. Price checks and inventory reviews (f) work on observed files only, without model calls. Each tool states what it measures, on whom and at what cost, so that a decision-maker knows what a result rests on.

\section{The trade area}\label{app:town}

\begin{figure}[!tbp]
  \centering
  \includegraphics[width=\linewidth,height=0.74\textheight,keepaspectratio]{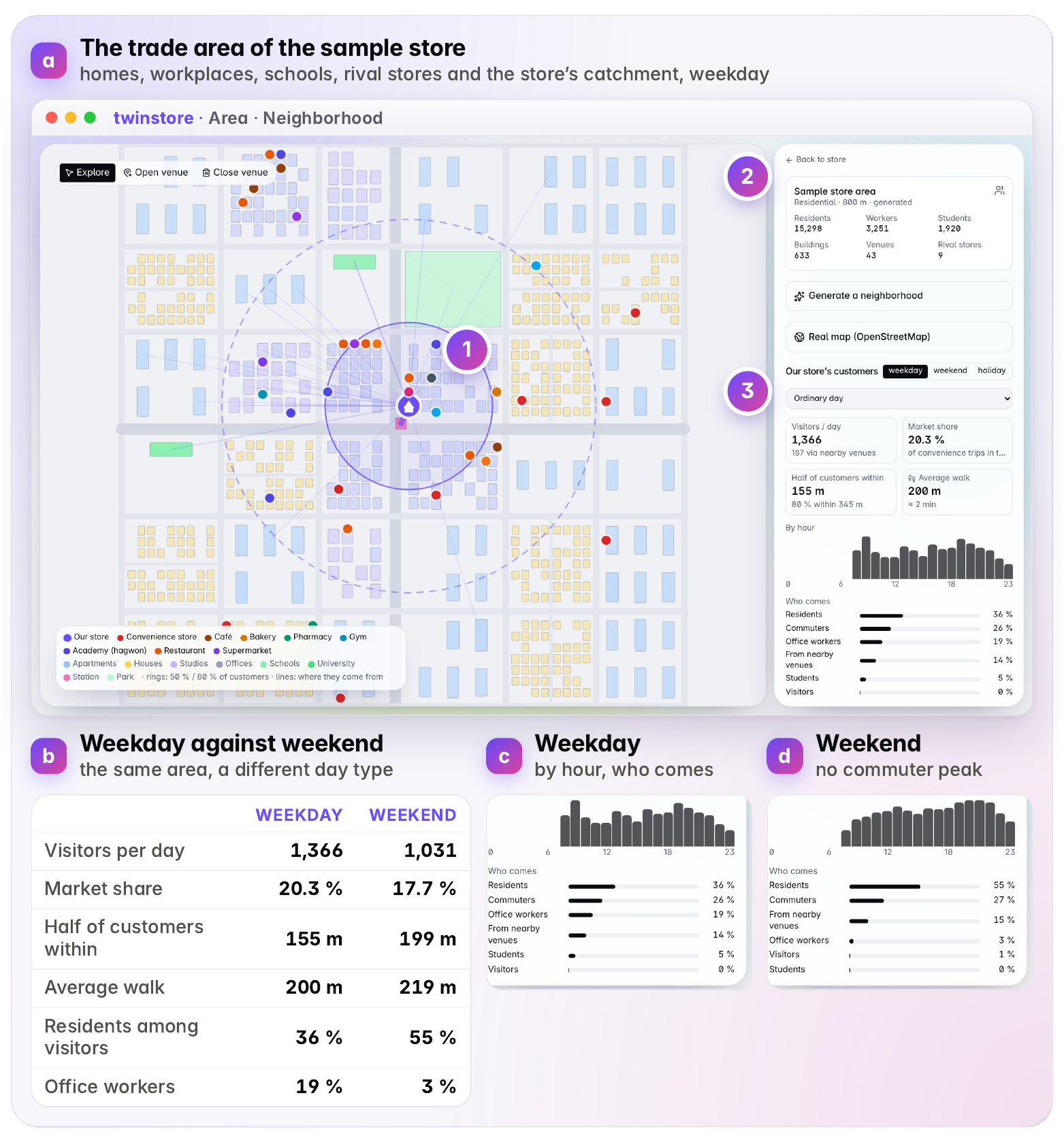}
  \caption{\textbf{The trade area of the sample store.} (a) The generated neighborhood with homes, workplaces, schools, venues and rival stores, the rings that contain 50\,\% and 80\,\% of the store's customers and the lines from where they come (1); the area's population (2) and the store's weekday visitors (3). (b) Weekday against weekend. (c, d) Visits by hour and who comes, on a weekday and at the weekend.}
  \label{fig:appx-town}
\end{figure}

The trade area of the sample store (\cref{fig:appx-town}a) is a generated residential neighborhood within 800\,m of the store: 633 buildings with 15,298 residents, 3,251 workers and 1,920 students, 43 venues and 9 rival stores (marker~2). The map draws homes, offices, schools, venues and rival stores, the store's catchment as two rings that contain 50\,\% and 80\,\% of its customers, and lines from where its customers come (1). Visits are allocated by a gravity model, and the panel reports the result for the chosen day type (3): on a weekday, 1,366 visitors a day, 187 of them on linked trips from nearby venues, a 20.3\,\% share of the area's convenience trips, half of the customers within 155\,m and an average walk of 200\,m.

Switching to the weekend (b--d) changes the population, not just the count. Visitors fall to 1,031 and the share to 17.7\,\%; customers come from farther, with half within 199\,m and an average walk of 219\,m; the early-morning peak disappears and visits build toward the evening; residents rise from 36\,\% to 55\,\% of visitors and office workers fall from 19\,\% to 3\,\%. These visitors can be sent into the store as the day's shoppers, each knowing where they came from (\cref{app:run}). For condition C1 the point is that the twin includes who is near the store and when, not only what is in it: a promotion aimed at commuters at 8\,am is evaluated on commuters at 8\,am.

\section{Working with the platform}\label{app:ops}

\begin{figure}[!tbp]
  \centering
  \includegraphics[width=\linewidth,height=0.74\textheight,keepaspectratio]{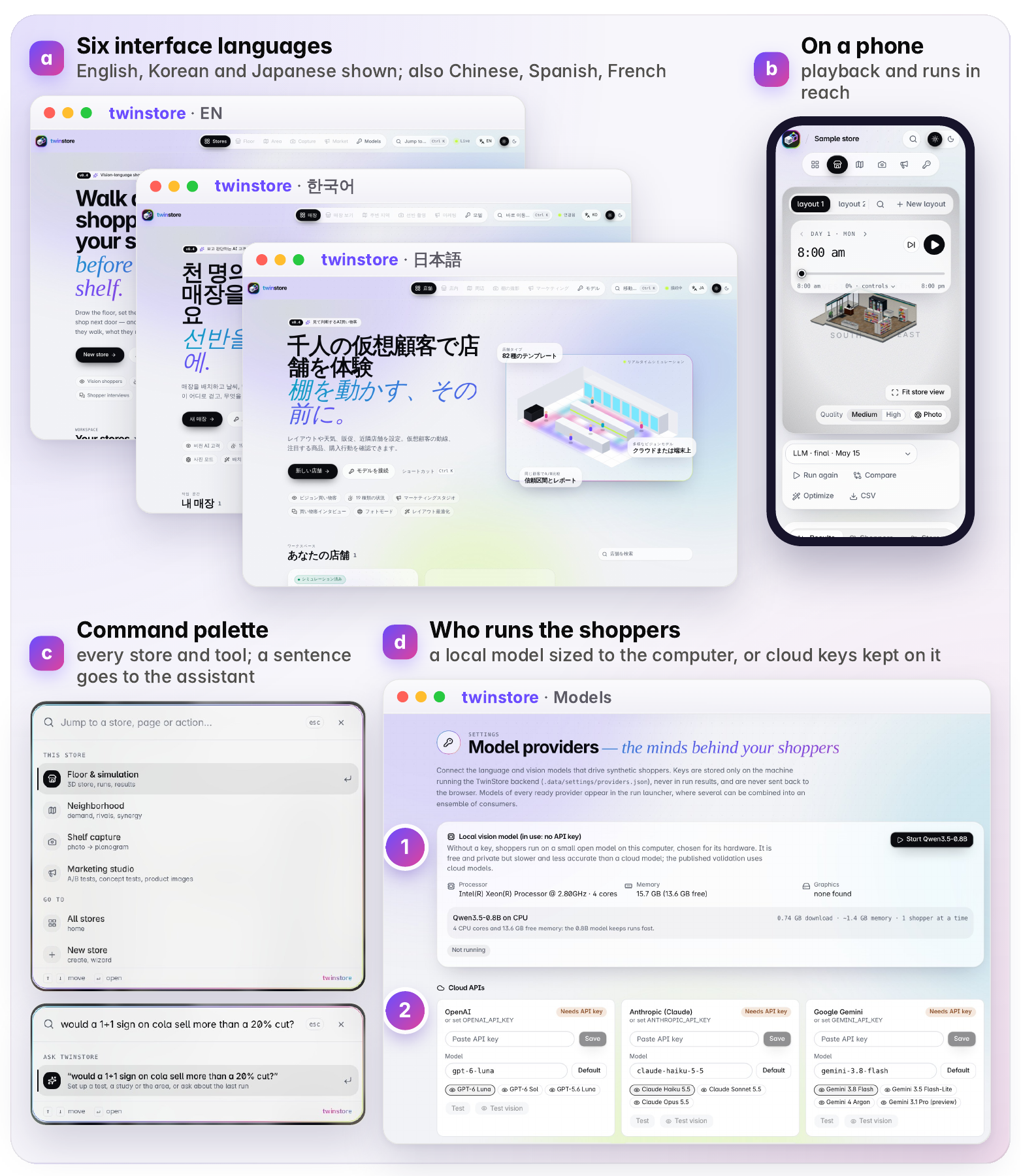}
  \caption{\textbf{Working with the platform.} (a) Three of the six interface languages. (b) The floor on a phone. (c) The command palette, which reaches every store and tool and passes a typed question to the assistant. (d) The model page: a local vision model chosen for the computer's hardware (1) and cloud providers whose keys stay on the machine (2).}
  \label{fig:appx-ops}
\end{figure}

The interface is available in six languages (\cref{fig:appx-ops}a) and adapts to phones and tablets, with playback and run controls kept within reach (b). A command palette (c) reaches every store, page and tool from the keyboard. A sentence typed into it, such as ``would a 1+1 sign on cola sell more than a 20\,\% cut?'', is offered to the assistant, which routes it to the tool that can answer it and opens a draft built from the same words; nothing runs until the user has checked the draft. The model page (d) decides who runs the shoppers. Without a key, the platform inspects the computer and proposes the open vision model that fits it (1), which then runs locally and privately. Keys for cloud providers are stored only on the machine that runs the engine, never in run results, and are not sent back to the browser (2).

These screens bear on design goal G4, use without a programmer, and on a practical condition for the argument of this appendix: checks that require a programmer are checks that few users will make. Every check in \cref{tab:appx-map} is a click, a switch or an upload.

\section{From the screens to the validation}\label{app:limits}

The screens and the validation answer complementary questions. \Cref{sec:validation} measures how close the receipts come to real ones on data the shoppers never saw. The screens show how those receipts are made, from the twin and the images a vision shopper receives to the trip, the results and the paired comparison, and give every user of the platform the means to look, on their own store, at why a result came out as it did.

\end{document}